\documentclass{article}
\PassOptionsToPackage{numbers}{natbib}
\usepackage[main, final]{neurips_2026}

\usepackage[utf8]{inputenc} 
\usepackage[T1]{fontenc}    
\usepackage{anyfontsize}
\usepackage{hyperref}       
\usepackage{url}            
\usepackage{booktabs}       
\usepackage{amsfonts}       
\usepackage{amsmath}
\usepackage{nicefrac}       
\usepackage{microtype}      
\usepackage{lipsum}
\usepackage{graphicx}
\usepackage{hyperref}
\usepackage{listings}
\usepackage{float}
\usepackage{xcolor}
\usepackage{soul}
\usepackage{makecell}
\usepackage{multirow}
\usepackage{enumitem}
\usepackage{tcolorbox}
\usepackage{algorithm}
\usepackage{algpseudocode}
\usepackage{xparse}
\usepackage{tikz}
\usetikzlibrary{positioning, arrows.meta, shapes, fit, backgrounds, calc, decorations.pathreplacing}
\usepackage{pgfplots}
\pgfplotsset{compat=1.18}
\usepackage{subcaption}
\usepackage{pifont}      
\usepackage{colortbl}    
\newcommand{\bx}{\mathbf{x}}
\usepackage{bm}

\usepackage{caption}[small,labelfont=bf]
\newcommand{\by}{\bm{y}}

\newcommand{\bM}{{\tt \bm{M}}}

\tcbuselibrary{breakable}
\tcbuselibrary{skins}
\graphicspath{ {./images/} }
\renewcommand{\arraystretch}{1.5}
\usepackage{bm}

\newcommand{\bg}{\bm{g}}
\definecolor{lightblue}{rgb}{0.678, 0.847, 0.902}
\definecolor{lightgreen}{rgb}{0.85,1,0.85}
\definecolor{lightred}{rgb}{1,0.85,0.85}
\definecolor{lightyellow}{rgb}{1,1,0.7}
\definecolor{darkgreen}{rgb}{0,0.5,0}
\definecolor{darkred}{rgb}{0.5,0,0}
\newcommand{\hlg}[1]{%
    {%
    \sethlcolor{lightgreen}%
    \hl{#1}%
    }%
}

\newcommand{\hlr}[1]{%
    {%
    \sethlcolor{lightred}%
    \hl{#1}%
    }%
}

\newcommand{\hlly}[1]{%
    {%
    \sethlcolor{lightyellow}%
    \hl{#1}%
    }%
}

\NewDocumentCommand\errsec{ s m }
{%
  \textbf{\color{blue}{%
      \IfBooleanTF{#1}{}{Section }%
      #2%
   } }%
}

\newcommand{\old}[1]{\ignorespaces}

\tcbset{
    emptyleaf/.style={
        enhanced,
        top=3pt-\baselineskip
    },
    basecategorytree/.style={
        width=\linewidth,
        sharp corners,
        boxsep=3pt,
        left=0pt,
        right=0pt,
        top=0pt,
        bottom=0pt,
        fonttitle={\small\fontfamily{put}\selectfont}
    },
    categorytree/.style={
        basecategorytree,
        width=0.95\linewidth,
        colback=#1!80,
        colframe=#1!80,
    },
    subcategorytree/.style={
        basecategorytree,
        colback=#1!70,
        colframe=#1!70,
    },
    subsubcategorytree/.style={
        basecategorytree,
        colback=#1!60,
        colframe=#1!60,
    },
    subsubsubcategorytree/.style={
        basecategorytree,
        colback=#1!50,
        colframe=#1!50,
    },
    subsubsubsubcategorytree/.style={
        basecategorytree,
        colback=#1!40,
        colframe=#1!40,
    },
}

\algrenewcommand\algorithmicindent{0.75em}
\algnewcommand\algorithmicforeach{\textbf{for each}}
\algdef{S}[FOR]{ForEach}[1]{\algorithmicforeach\ #1\ \algorithmicdo}

\renewcommand\algorithmicdo{}

\usepackage{wrapfig}

\usepackage{xcolor}

\title{Constrained Code Generation with Diffusion Language Models}
\author{%
  Lize Shao\thanks{Equal contribution.}\quad
  Michael Cardei\footnotemark[1]\quad
  Zichen Xie\quad
  Ferdinando Fioretto\thanks{Equal senior-author contribution.}\quad
  Wenxi Wang\footnotemark[2] \\
  University of Virginia \\
  \texttt{\{zgr3et, ntr2rm, graysonxie, fioretto, wenxiw\}@virginia.edu} \\
}
\begin{document}

\maketitle

\begin{abstract}
Diffusion language models (DLMs) are a powerful, emerging paradigm for code generation. They construct programs through iterative refinement of partially corrupted token sequences and enable parallel token refinement. Importantly, this paradigm exposes a global program state at each denoising step, which provides a natural intervention point for enforcing program-level functionality and security constraints, guiding the generation before the final code is committed. Building on this observation, the paper introduces \textit{Constrained Diffusion for Code} (CDC), an inference-time neurosymbolic framework that integrates constraint satisfaction directly into the reverse denoising process without retraining the base diffusion model. CDC augments the base diffusion sampler with constraint-aware denoising operators that combine mathematical optimization with program analysis to identify constraint-relevant regions of the intermediate program state and locally adjust the denoising trajectory, steering generation toward feasible programs while remaining close to the base  model. Across code generation benchmarks, CDC consistently improves constraint satisfaction in functional correctness, security, and even syntax, outperforming diffusion and autoregressive baselines with less corrective computation and more localized edits. 
\begin{figure}[h]
    \vspace{-4pt}
    \centering
    \scalebox{0.95}{\includegraphics[width=1.00\linewidth,height=150pt,trim=20 20 20 20]{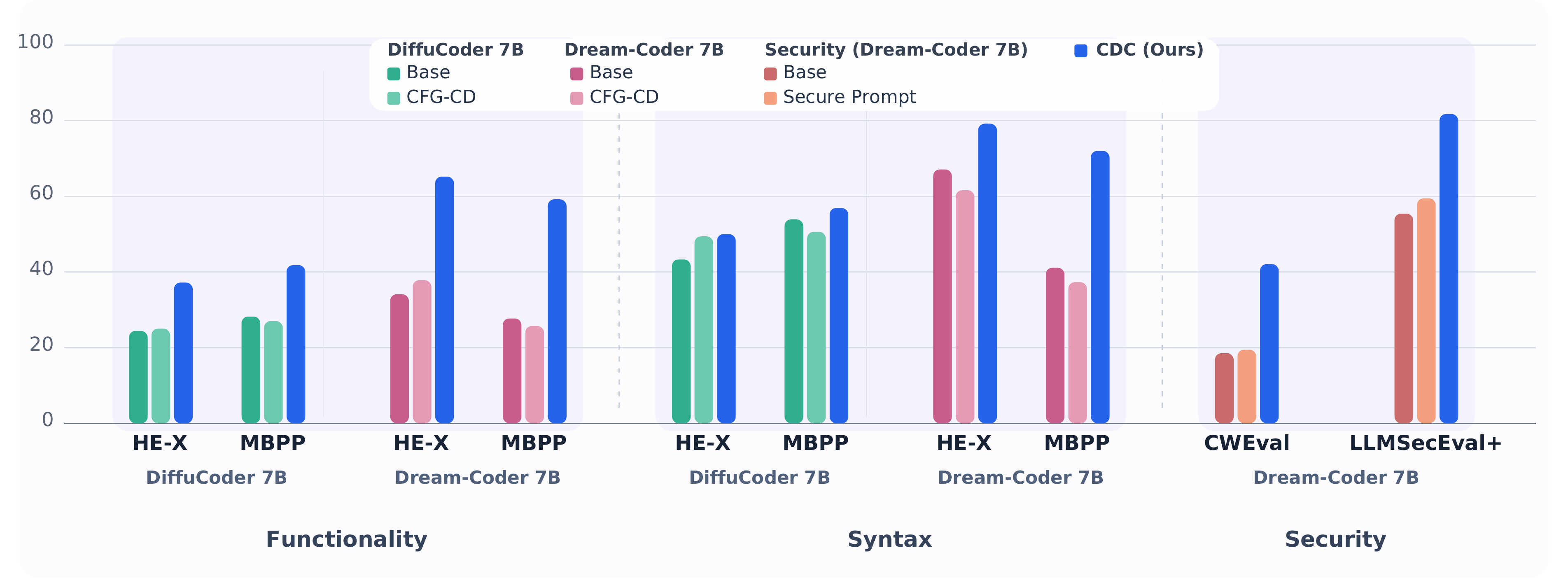}\
    }
    \vspace{-8pt}
    \caption{\small CDC vs.~other diffusion code baselines on HumanEval-X (HE-X), MBPP, CWEval and LLMSecEval+.}
    \vspace{-22pt}
    \label{fig:strong_results}
\end{figure}
\end{abstract}










\vspace{-1ex}
\section{Introduction}
\vspace{-1ex}
Code generation has become an increasingly important capability in modern generative modeling systems, with notable applications in program synthesis, developer assistance, and agentic software workflows \cite{li2022alphacode, zhang2023repocoderrepositorylevelcodecompletion, yang2024sweagentagentcomputerinterfacesenable}. 
In this setting, diffusion language models (DLMs) are emerging as a highly promising paradigm, constructing programs through iterative refinement of partially corrupted token sequences \cite{gong2025diffucoder, xie2025dreamcoder}. 
However, the objective used to train generative models does not by itself capture the requirements that make generated code correct or secure. These models are trained to match a target distribution and produce high-fidelity samples, but a program may be highly plausible under the base model while still producing incorrect behavior, or containing an exploitable vulnerability. In practice, generated code must satisfy explicit program-level constraints, including syntactic validity and, more importantly, functional correctness and security properties. Since programs are executable objects, violations of these constraints can lead directly to runtime errors, incorrect outputs, or security failures \cite{pearce2021asleepkeyboardassessingsecurity, basic2026vulnerabilitiesremediationsystematicliterature}. Incorporating such constraints is therefore critical for both code quality and safety.

These requirements expose a central challenge for code generation: \emph{Program-level constraints often depend on global structure or semantic properties that are not visible from any single local token decision}. More broadly, common guidance mechanisms such as prompting, reranking, rejection sampling, or post-hoc repair typically intervene only after candidate programs have already been generated \cite{tony2025promptingtechniquessecurecode, chen2023teachinglargelanguagemodels}. In code generation, this is particularly limiting because the underlying search space is highly combinatorial, while the subset of programs satisfying semantic or security constraints may constitute only a small feasible region under the model distribution. Thus, post-hoc approaches can be both computationally inefficient and unreliable, especially when constraint satisfaction requires steering generation away from the model's dominant unconstrained modes and toward rarer but feasible solutions. In contrast, as this paper shows, diffusion language models offer a different opportunity: their approach to refine the entire sequence through intermediate denoising states exposes a global program representation throughout sampling. This structure creates natural intervention points for integrating constraints directly into the generated code during the denoising process. 

Building on this opportunity, this paper introduces \textit{Constrained Diffusion for Code} (CDC), an inference-time neurosymbolic framework that integrates program-level constraints (i.e., functionality and security) directly into the reverse denoising process of diffusion language models, without retraining the base model. CDC treats diffusion sampling as an editable program trajectory: at each denoising step, the model proposes a full clean-state distribution over the program; CDC decodes this proposal, evaluates the resulting candidate against program-level constraints, localizes feedback to relevant regions, and applies a constraint-aware correction before sampling continues. Each correction follows a localized constrained-sampling principle: reduce constraint violation while remaining close to the base denoiser, revising constraint-relevant regions without unnecessarily perturbing the rest of the program. The framework supports both soft and hard feedback. For functional correctness, CDC instantiates this principle as GradGuide, which uses soft surrogate execution signals to localize likely errors and guide targeted denoising updates. For security, CDC instantiates it as MDFI, which uses static program analysis to identify vulnerable regions and correct them through localized remasking, insertion, and feedback injection. In this way, CDC exploits the global, editable states exposed by diffusion sampling to perform focused constraint correction during generation, rather than relying on post-hoc filtering or whole-program regeneration after failure.

This work makes three main contributions:
\begin{enumerate}[nosep, leftmargin=*]
    \item \textbf{Constraint-aware localization.} It introduces a general localization mechanism for constrained diffusion code generation, mapping program-level feedback from surrogates or static analyzers to the token regions that should be revised during denoising.

    \item \textbf{Constraint-aware correction.} It proposes inference-time correction operators that revise these localized regions during reverse diffusion while staying close to the base denoiser. The operators support differentiable surrogate guidance through gradient and KL-anchored updates, and non-differentiable symbolic analyzer feedback through targeted remasking, mask insertion, and feedback injection.

    \item \textbf{Empirical evaluation.} It provides empirical evidence that denoising-time constraint integration improves constrained code generation across functional-correctness and security-oriented benchmarks. 
    As summarized in Figure~\ref{fig:strong_results}, CDC improves functionality from 34.1\% to 65.2\% on HumanEval-X C++ and from 27.7\% to 59.2\% on MBPP-C++ over Dream-Coder 7B, while MDFI improves CWEval joint functionality-security success from 12.0\% to 34.3\% over the state-of-the-art diffusion baseline with far few edited tokens. Although CDC targets functionality and security, it also improves syntactic correctness from 67.1\% to 79.3\% on HumanEval-X C++ and from 41.1\% to 72.0\% on MBPP-C++.
\end{enumerate}

\section{Related Work}
\vspace{-1ex}
Discrete diffusion language models generate sequences through iterative denoising rather than left-to-right decoding, exposing partially generated global states throughout sampling. Recent code-oriented diffusion models including DiffuCoder~\cite{gong2025diffucoder} and Dream-Coder 7B~\cite{xie2025dreamcoder} show that this paradigm can achieve competitive code-generation performance while supporting flexible, non-autoregressive refinement. CDC builds on this property: rather than using diffusion only as an alternative decoder, it treats intermediate denoising states as editable program representations where constraints can be evaluated, localized, and incorporated before generation is complete.

Controllable code generation has been studied primarily in autoregressive models through constrained decoding, reranking, rejection sampling, verification, and iterative repair~\cite{hu-etal-2019-improved, ni2023leverlearningverifylanguagetocode, chen2023teachinglargelanguagemodels, M_ndler_2025, bhatt2024cyberseceval2widerangingcybersecurity}. These methods can improve syntactic validity, type consistency, functional correctness, or security, but they typically operate either at the next-token level or after a complete candidate has been generated. This is limiting for program-level constraints: semantic and security violations often depend on global dataflow or execution behavior, and repairing them post hoc may require regenerating large suffixes or entire programs. CDC instead injects feedback during denoising, when the full program state is visible and localized regions can still be revised.

Recent work has explored controllable generation for discrete diffusion models through classifier guidance, classifier-free guidance, hidden-state optimization, and projection-based constrained sampling~\cite{schiff2025simpleguidancemechanismsdiscrete, gruver2023proteindesignguideddiscrete, cardei2025constraineddiscretediffusion, christopher2025neurosymbolicgenerativediffusionmodels}. Closest to our setting, M\"undler et al.~\cite{muendler2025cfgdiffusion} enforce context-free grammar constraints during diffusion-LLM decoding, improving syntactic correctness for code. CDC targets a broader class of program-level constraints, including functional and security requirements that cannot generally be expressed as grammar membership or local token restrictions. Rather than applying global guidance or syntax-only filtering, CDC localizes surrogate or analyzer feedback to constraint-relevant code regions and revises those regions during the reverse diffusion trajectory.

\vspace{-1ex}
\section{Problem Setting}
\vspace{-1ex}
\label{subsec:problem_setting}
\noindent\textbf{Code generation with target specifications.}
Let $\bx_0 = (x_0^1,\dots,x_0^L)$ denote a length-$L$ sequence of discrete tokens, where each token takes values in a vocabulary $\mathcal{V}$. In code generation, $\mathcal{V}$ may include programming-language keywords, identifiers, literals, punctuation, operators, and special formatting tokens (e.g., \texttt{def}, \texttt{if}, \texttt{=}, \texttt{(}, \texttt{)}, \texttt{:}, variable names). A code generation model defines a conditional distribution over token sequences given a context $c$, where the functionality or security requirements may be specified through a problem description, function signature, partial program, unit tests, or other task-specific information. We refer to these requirements as the \emph{target specification}; Figure~\ref{fig:methods} shows an example.  The goal is therefore to generate a sequence $\bx_0$ that is both plausible under the learned code distribution and valid with respect to the target specification.

\noindent\textbf{Constrained sampling with hard and soft constraints.}
Let $p_\theta(\bx \mid c)$ denote a base code generation model. Unconstrained decoding samples likely programs from $p_\theta$; constrained decoding instead defines a sampler $p_\theta^{\mathcal{C}}(\bx \mid c)$ that remains close to the base model while steering generation toward the feasible set $\mathcal{C}(c)$ specified by the task.
We distinguish two kinds of constraints. \emph{Hard constraints} are feasibility conditions that must hold exactly, such as parsing, compiling, passing required tests, or satisfying security checks. \emph{Soft constraints} provide graded feedback before exact feasibility is reached, such as surrogate correctness scores, partial test-pass rates, or analyzer severity scores. CDC uses both: hard constraints define the target feasible set, while soft constraints provide useful denoising-time guidance toward it.

\vspace{-1ex}
\section{Preliminaries: Masked Diffusion Models}
\vspace{-1ex}
Masked diffusion language models generate a sequence by reversing a discrete noising process \cite{ye2025dream, sahoo2024simpleeffectivemaskeddiffusion, NEURIPS2024_bad233b9, gong2025diffucoder}. Starting from a clean token sequence $\bx_0$, the forward process gradually corrupts tokens into an absorbing mask token $\texttt{[MASK]}$; once a token is masked, it remains masked at later timesteps \cite{sahoo2024simpleeffectivemaskeddiffusion,nie2025llada}.
For each position $i \in [L]$, the forward transition is
\begin{equation}
\label{eq:forward_transition}
q(x_t^i \mid x_{t-1}^i)
=
\mathrm{Cat}\!\left(
x_t^i;\;
\alpha_{t\mid t-1}x_{t-1}^i + \bigl(1-\alpha_{t\mid t-1}\bigr)\bM
\right),
\end{equation}
where $\mathrm{Cat}(\cdot;\pi)$ is a categorical distribution, $\bM$ is the one-hot representation of $\texttt{[MASK]}$, and $\alpha_{t\mid t-1}:=\alpha_t/\alpha_{t-1}$ for a decreasing noise schedule with $\alpha_0=1$ and $\alpha_T=0$. Thus, an unmasked token is preserved with probability $\alpha_{t\mid t-1}$ and masked otherwise.

The reverse process uses a neural denoiser to fill in masked positions. Given a partially masked sequence $\bx_t$ at timestep $t$, the denoiser predicts a clean-token distribution
$\hat{\bx}_0^{(t)}=\bx_\theta(\bx_t,t)$.
The reverse transition is
\begin{equation}
\label{eq:reverse_kernel}
p_\theta(x_{t-1}^i \mid \bx_t)
=
\begin{cases}
\mathrm{Cat}(x_{t-1}^i; x_t^i), & \text{if } x_t^i \neq \texttt{[MASK]}, \\[0.8em]
\mathrm{Cat}\!\left(
x_{t-1}^i;\;
\gamma_t \bM + \eta_t \bx_\theta^{i}(\bx_t,t)
\right), & \text{if } x_t^i = \texttt{[MASK]},
\end{cases}
\end{equation}
for each position $i \in [L]$ and timestep $t=T,\dots,1$, with
$
\gamma_t = \frac{1-\alpha_{t-1}}{1-\alpha_t},
\eta_t = \frac{\alpha_{t-1}-\alpha_t}{1-\alpha_t}
$
balancing mask retention and sampling from the denoiser's clean-token prediction.

At inference time, sampling starts from the fully masked sequence $\bx_T$ and repeatedly applies Eq.~\ref{eq:reverse_kernel} until it obtains the generated sequence $\bx_0$. 

\vspace{-1ex}
\section{Constrained Diffusion for Code Generation}
\vspace{-1ex}
\label{sec:method}

\begin{figure}[t]
    \centering
    \includegraphics[width=1.00\linewidth]{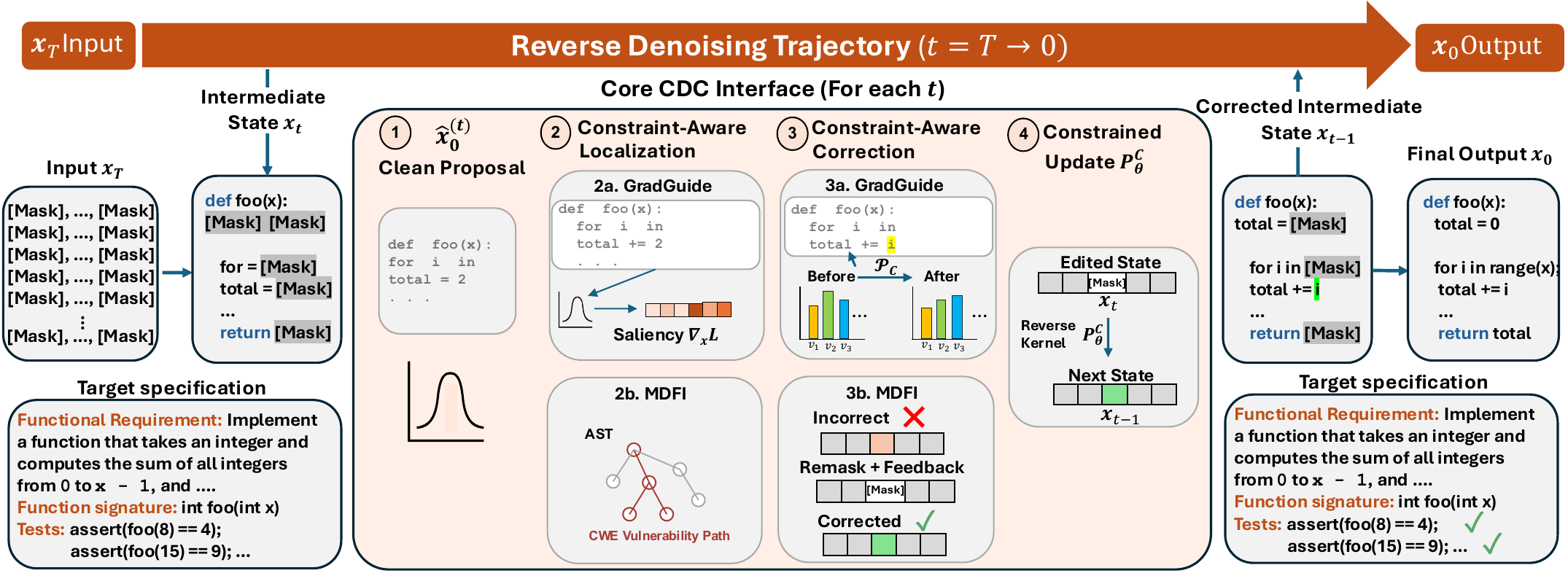}
    \vspace{-10pt}
    \caption{Overview of CDC. }
    \vspace{-15pt}
    \label{fig:methods}
\end{figure}

The formulation above motivates \emph{Constrained Diffusion for Code} (CDC): at each timestep, the denoiser proposes a full clean program distribution $\hat{\bx}_0^{(t)}$, creating a natural point to evaluate, localize, and incorporate program-level constraints during reverse denoising.
\vspace{-1ex}
\subsection{CDC Framework}
\vspace{-1ex}
\label{subsec:constrained_denoising}
CDC preserves the base diffusion model without additional training and formulates reverse diffusion as an editable trajectory that integrates base denoising with constraint-aware interventions. Starting from a fully masked program $\bx_T \!=\! ({\small\texttt{[MASK]}},\dots,$ ${\small \texttt{[MASK]}})$, sampling proceeds backward through timesteps $t=T,\dots,1$ until a clean program $\bx_0$ is produced. The overview of CDC is shown in Figure~\ref{fig:methods}. At each timestep, CDC performs four steps:

\textbf{Step 1: Clean-state proposal.}
Given the current partially masked sequence $\bx_t$, the base denoiser predicts a per-position categorical distribution over clean tokens,
\begin{equation}
\label{eq:clean_state_proposal}
    \hat{\bx}_0^{(t)}
    =
    \bx_\theta(\bx_t,t)
    \in
    \Delta^{L \times |\mathcal{V}|}.
\end{equation}
This distribution represents the model's unconstrained clean-state proposal at timestep $t$, which can be decoded into an intermediate program candidate for constraint evaluation, for example by taking the per-position maximum,
\begin{equation}
\label{eq:decode_intermediate}
    \bar{x}_{0}^{(t),i}
    =
    \arg\max_{v \in \mathcal{V}}
    \hat{x}_{0}^{(t),i}(v),
    \qquad i \in [L].
\end{equation}

\textbf{Step 2: Constraint-aware localization.}
A constraint evaluator $\mathcal{E}$ then evaluates this candidate $\bar{\bx}_0^{(t)}$ under the task context $c$ and constraint specification $\mathcal{C}$. The evaluator $\mathcal{E}$ may be instantiated by any constraint-scoring mechanism, such as a symbolic program analyzer, a verifier, or a learned surrogate model. The evaluator returns violation scores $\boldsymbol{\nu}_t$ and structured feedback $r_t$ (e.g., a vulnerability report), which are mapped by a constraint-localization map $\mathcal{M}$ to editable regions $\mathcal{S}_t$:

\begin{equation}
\label{eq:constraint_localization}
    (\boldsymbol{\nu}_t,r_t)
    =
    \mathcal{E}(\bar{\bx}_0^{(t)}\!\!,c;\;\mathcal{C}),
    \qquad
    \mathcal{S}_t
    =
    \mathcal{M}(\bar{\bx}_0^{(t)}\!\!,c,r_t;\;\mathcal{C})
    \subseteq [L].
\end{equation}

The map $\mathcal{M}$ is induced by the form of feedback returned by $\mathcal{E}$, and translates evaluator feedback into token positions or spans that are responsible for the constraint violation. The set $\mathcal{S}_t$ identifies where the denoising trajectory should be revised, with specific instantiations given in Section~\ref{subsec:operator}.
This separates the question of \emph{where} to edit from the question of \emph{how} to modify the denoising trajectory, which is the focus of the next step.

\textbf{Step 3: Constraint-aware correction.}
Given the localized edit region $\mathcal{S}_t$ and feedback $r_t$, CDC applies a constraint-aware denoising operator $\mathcal{P}_{\mathcal{C}}$ to produce a corrected clean-state proposal:
\begin{equation}
\label{eq:operator_application}
    \by_t
    =
    \mathcal{P}_{\mathcal{C}}
    (
        \hat{\bx}_0^{(t)},
        \mathcal{S}_t,
        r_t,
        c
    ).
\end{equation}

The operator is designed to reduce constraint violation while keeping $\by_t$ close to the base denoiser output $\hat{\bx}_0^{(t)}$. CDC represents constraint feedback through violation penalties that may come from differentiable surrogate scores, verifier outputs, or security diagnostics. For constraint $j$ at timestep $t$, let $\nu_{t,j}(\by; r_t,c,\mathcal{S}_t)$ denote the feedback-induced penalty under candidate proposal $\by$, and let $\lambda_j \geq 0$ denote its penalty coefficient. The resulting localized trust-region update is:
\begin{equation}
\label{eq:constraint_operator}
\by_t = 
\arg\min_{\by \in \Delta^{L \times |\mathcal{V}|}}
\Bigl\{
D_{\mathrm{KL}}\!\left(
\by
\,\|\,
\hat{\bx}_{0}^{(t)}
\right)
+
\underbrace{\sum_{j=1}^{m} \lambda_j \, \nu_{t,j}(\by; r_t,c,\mathcal{S}_t)}_{V_t(\by; r_t,c,\mathcal{S}_t)}
\Bigr\}.
\end{equation}
Here, $D_{\mathrm{KL}}$ denotes Kullback--Leibler divergence, which anchors the correction to the base proposal, and $V_t$ is the aggregate violation penalty. The localization set $\mathcal{S}_t$ scopes the correction: some operators enforce locality by anchoring positions outside $\mathcal{S}_t$, while others use $\mathcal{S}_t$ as a remasking target or feedback anchor. Section~\ref{subsec:operator} instantiates this interface for differentiable surrogate feedback for program functionality, and symbolic program-analysis feedback for program security.

\textbf{Step 4: Constrained reverse update.}
Finally, CDC advances the reverse chain using the corrected proposal $\by_t$ through a constraint-aware reverse kernel,
\begin{equation}
\label{eq:constrained_reverse_update}
    \bx_{t-1}
    \sim
    p_\theta^{\mathcal{C}}(\cdot \mid \bx_t,\by_t).
\end{equation}
A representative transition replaces the base clean-token distribution prediction with the corrected proposal:
\begin{equation}
\label{eq:constrained_reverse_kernel}
p_\theta^{\mathcal{C}}(x_{t-1}^i \mid \bx_t,\by_t)
=
\begin{cases}
\mathrm{Cat}(x_{t-1}^i; x_t^i),
& \text{if } x_t^i \neq \texttt{[MASK]}, \\[0.8em]
\mathrm{Cat}\!\left(
x_{t-1}^i;\;
\gamma_t \bM + \eta_t \by_t^i
\right),
& \text{if } x_t^i = \texttt{[MASK]}.
\end{cases}
\end{equation}
Here, $\by_t^i$ is the corrected clean-token distribution at position $i$. 
Depending on the operator, positions outside $\mathcal{S}_t$ may either remain anchored to the base proposal
or receive weaker global adjustments, while positions inside $\mathcal{S}_t$ receive the strongest constraint-aware
corrections through modified token distributions, targeted remasking, or feedback-conditioned revision,
as described in Section~\ref{subsec:operator}. The resulting reverse trajectory is illustrated below.
\begin{tcolorbox}[
  enhanced,
  boxrule=0.6pt, colback=gray!8, colframe=gray!60, arc=2mm,
  boxsep=0.5pt,
  left=0pt, right=0pt,
  top=2pt, bottom=2pt
]
\begin{equation*}
\bx_t
\xrightarrow{\;\bx_\theta(\cdot,t)\;}
\hat{\bx}_0^{(t)}
\xrightarrow{\;\mathrm{Decode}\;}
\bar{\bx}_0^{(t)}
\xrightarrow{\;\mathcal{E},\,\mathcal{M}\;}
(\boldsymbol{\nu}_t,r_t,\mathcal{S}_t)
\xrightarrow{\;\mathcal{P}_{\mathcal{C}}\;}
\by_t
\xrightarrow{\;p_\theta^{\mathcal{C}}\;}
\bx_{t-1}.
\end{equation*}
\end{tcolorbox}

\subsection{Operator Instantiations}
\label{subsec:operator}

We now show how CDC becomes concrete by instantiating the three components of the framework: evaluator $\mathcal{E}$, localization map $\mathcal{M}$, and correction operator $\mathcal{P}_{\mathcal{C}}$. Following the problem setting (Section~\ref{subsec:problem_setting}), hard constraints define feasibility, while soft constraints provide graded signals that can guide the sampler before feasibility is reached. Our first instantiation, GradGuide, uses this soft-feedback route for program functionality: a learned surrogate relaxes hard execution outcomes into soft scores, localizes likely functional errors with gradients, and corrects the localized region through KL-anchored proposal updates and remasking. Our second instantiation, Mid-Diffusion Feedback Injection (MDFI), uses the hard-feedback route for program security: a static analyzer reports concrete vulnerability witnesses on partial programs, localizes them on a program graph, and corrects the corresponding regions through targeted remasking and feedback injection.

\textbf{Surrogate-gradient operator: GradGuide.}
Functional correctness is ultimately a hard constraint: the final program must pass the required tests. During denoising, however, this hard constraint is not differentiable, so GradGuide instantiates the evaluator $\mathcal{E}^{\mathrm{GG}}$ with an auxiliary surrogate $g_\phi$ trained ahead of time with execution-driven labels (e.g., test-pass outcomes), while keeping the diffusion model parameters $\theta$ fixed. 
Given the clean-state proposal $\hat{\bx}_0^{(t)}$, GradGuide maps each predicted token distribution to an expected embedding,
\begin{equation}
\label{eq:gg-embedding}
    \mathrm{Emb}(\hat{\bx}_0^{(t)})^i
    =
    \sum_{v\in\mathcal{V}}
    \hat{x}_0^{(t),i}(v)\mathbf{e}_v
    =
    \hat{x}_0^{(t),i}\mathbf{E}_{\mathrm{tok}},
    \qquad i\in[L],
\end{equation}
where $\mathbf{E}_{\mathrm{tok}}$ is the token-embedding matrix, and $\mathbf{e}_v$ denotes the embedding of token $v$.

The surrogate predicts a satisfaction score  ${\nu}^{\mathrm{GG}}_{\phi,j}(\mathrm{Emb}(\hat{\bx}_0^{(t)}),c)$ for each constraint $j$, which is converted into the relaxed violation
\begin{equation}
\label{eq:gg-violation}
    \Delta \nu^{\mathrm{GG}}_j(\hat{\bx}_0^{(t)},c)
    =
    \max\!\left(
    0,\,
    \tau_j -
    \nu^{\mathrm{GG}}_{\phi,j}(\mathrm{Emb}(\hat{\bx}_0^{(t)}),c)
    \right),
    \qquad
     V^{\mathrm{GG}}_t
    =
    \sum_{j=1}^{m}
    \Delta \nu^{\mathrm{GG}}_j .
\end{equation}
GradGuide localizes edits with one backward pass through $g_\phi$. The saliency of position $i$ is
\begin{equation}
\label{eq:gg-saliency}
    a_i
    =
    \left\|
    \nabla_{\mathrm{Emb}(\hat{\bx}_0^{(t)})^i}
V^{\mathrm{GG}}_t(\hat{\bx}_0^{(t)},c)
    \right\|_2,
    \qquad
    \mathcal{S}_t^{\mathrm{GG}}
    =
    \mathrm{Expand}\!\Bigl(
    \mathrm{Top}\text{-}k(a_1,\dots,a_L)
    \Bigr),
\end{equation}
where $\mathrm{Expand}(\cdot)$ maps selected token positions to syntactically
coherent edit spans by enlarging each position to a local syntactic window. Thus, $\mathcal{S}_t^{\mathrm{GG}}$ contains the syntactically expanded token spans that most influence the surrogate-predicted violation.

Given $\mathcal{S}_t^{\mathrm{GG}}$, GradGuide computes a corrected proposal by approximately minimizing Eq.~\ref{eq:constraint_operator} with $V_t$ replaced by $V^{\mathrm{GG}}_t$.  

The resulting $\by_t$ is used in the constrained reverse kernel. When the intermediate program is sufficiently decoded but still violates the hard functional constraint, GradGuide also reopens the localized region by setting $x_t^{i}=\texttt{[MASK]}$ for $i\in\mathcal{S}_t^{\mathrm{GG}}$, allowing the remaining denoising steps to regenerate those positions under the corrected proposal. Additional details, including surrogate training and other implementation details, are provided in Appendix~\ref{app:gradguide}.

\textbf{Program-analysis-guided operator: MDFI.}
Security requirements are hard constraints: a program is feasible only if the relevant vulnerability checks pass. For these constraints, we use a static program analyzer to instantiate the evaluator $\mathcal{E}$, which is applied to the intermediate proposal $\bar{\bx}_0^{(t)}$ at selected denoising checkpoints. The analyzer constructs a partial program graph $\mathcal{G}_t$ from the partial program structure and returns vulnerability witnesses
\begin{equation}
\label{eq:mdfi-witness}
    w_k=(n_k,\tau_k,h_k),
\end{equation}
where $n_k$ is the offending graph node, $\tau_k\in\{\mathrm{sub},\mathrm{ins}\}$ indicates whether the repair is substitution- or insertion-like, and $h_k$ is a remediation hint. The witness set $\mathcal{W}_t=\{w_k\}_{k=1}^{K_t}$ defines the structured feedback $r_t^{\mathrm{MDFI}}$ and the corresponding hard violation vector $\boldsymbol{\nu}_t^{\mathrm{MDFI}}$.

MDFI localizes each witness by taking its analyzer-supported abstract syntax tree and dataflow neighborhood and projecting it back to token positions:
\begin{equation}
\label{eq:mdfi-localization}
    \mathcal{S}_t^{\mathrm{MDFI}}
    =
    \mathrm{TopBudget}_{B}
    \!\left(
    \bigcup_{k=1}^{K_t}
    \mathrm{Tok}\bigl(\mathcal{N}(n_k;\mathcal{G}_t)\bigr)
    \right)
    \subseteq [L].
\end{equation}
Here, $\mathrm{TopBudget}_B(\cdot)$ selects top token spans under a token budget $B$, prioritizing higher-confidence analyzer witnesses, while $\mathcal{N}(n_k;\mathcal{G}_t)$ denotes the relevant syntactic or dataflow neighborhood around the witness, and $\mathrm{Tok}(\cdot)$ maps graph nodes back to token spans. Thus, $\mathcal{S}_t^{\mathrm{MDFI}}$ contains diagnostic-supported regions.

Given $\mathcal S_t^{\mathrm{MDFI}}$ and $r_t^{\mathrm{MDFI}}$, MDFI applies
three discrete interventions.  Let $\mathcal S_t^{\mathrm{sub}}$ and
$\mathcal S_t^{\mathrm{ins}}$ denote the portions of
$\mathcal S_t^{\mathrm{MDFI}}$ supported by substitution-type and insertion-type
witnesses, respectively:
\begin{equation}
\mathcal S_t^{a}
=
\mathcal S_t^{\mathrm{MDFI}}
\cap
\bigcup_{k:\tau_k=a}
\mathrm{Tok}\!\left(\mathcal N(n_k;\mathcal G_t)\right),
\qquad
a\in\{\mathrm{sub},\mathrm{ins}\}.
\end{equation}
For substitution-type witnesses, MDFI remasks the offending committed region:
\begin{equation}
x_t^{\star,i}
=
\begin{cases}
\texttt{[MASK]}, & i\in\mathcal S_t^{\mathrm{sub}},\\
x_t^i, & i\notin\mathcal S_t^{\mathrm{sub}}.
\end{cases}
\end{equation}
For insertion-type witnesses, MDFI inserts $K$ fresh \texttt{[MASK]} tokens near
a structural anchor selected from $\mathcal S_t^{\mathrm{ins}}$:
\begin{equation}
\mathbf x_t^\star
\leftarrow
\mathrm{Insert}_{K}
\bigl(
\mathbf x_t^\star,
\mathrm{anchor}(w_k;\mathcal S_t^{\mathrm{ins}})
\bigr),
\qquad
k:\tau_k=\mathrm{ins},
\end{equation}
where $\mathrm{Insert}_K(x,a)$ inserts $K$ \texttt{[MASK]} tokens into
sequence $x$ at anchor position $a$. The anchor function selects a syntactically
valid insertion point near the insertion-type witness.

The MDFI update is therefore
\begin{equation}
\label{eq:mdfi-step}
    \bigl(
    \bx_t^\star,\,
    c^\star,\,
    \mathcal{S}_t^{\mathrm{MDFI}}
    \bigr)
    =
    \mathcal{P}_{\mathcal{C}}^{\mathrm{MDFI}}
    \bigl(
    \hat{\bx}_0^{(t)},\,
    \mathcal{S}_t^{\mathrm{MDFI}},\,
    r_t^{\mathrm{MDFI}},\,
    c
    \bigr).
\end{equation}
Here, $\mathcal{P}_{\mathcal{C}}^{\mathrm{MDFI}}$ is the MDFI instantiation of the constraint-aware projection operator defined in Eq.~\ref{eq:operator_application} and operationalized by the  objective in Eq.~\ref{eq:constraint_operator}.
The chain continues with $\bx_t$ replaced by $\bx_t^\star$ and $c$ replaced by $c^\star$. MDFI does not update $\theta$ or run an inner re-denoising loop; it edits the current state/context in place and lets the remaining reverse steps regenerate the affected regions. If no checkpoint fires or no witness is detected, MDFI reduces to the identity operator. Additional details are in Appendix~\ref{app:mdfi}.

\vspace{-1ex}
\section{Experiments}
\vspace{-1ex}
\label{sec:experimental}
\begin{table}[t]
  \caption{Functional correctness (\%) on HumanEval-X C++ and MBPP-C++.  Parenthesized values report the absolute change from the Vanilla model: \hlg{green} indicates an improvement of $\geq$10\%, \hlly{yellow} an improvement of $<$10\%, and \hlr{red} a regression. Within each model, the best performance is \textbf{bolded}.}
  \label{tab:main_functional}
  \centering
  \renewcommand{\arraystretch}{1.05}
  \setlength{\tabcolsep}{4pt}
  \scalebox{0.7}{
  \begin{tabular}{@{}l l lll lll@{}}
  \toprule
  & & \multicolumn{3}{c}{\textbf{HumanEval-X C++}} & \multicolumn{3}{c}{\textbf{MBPP-C++}} \\
  \cmidrule(lr){3-5} \cmidrule(lr){6-8}
  \textbf{Model} & \textbf{Method} & Syn. (compile) & Fun. (p@1) & Fun. (p@10) & Syn. (compile) & Fun. (p@1) & Fun. (p@10) \\
  \midrule
  
  \multirow{3}{*}{Dream 7B}
    & Vanilla     & 40.2   & 10.4   & 51.8   & 60.7   & 25.4   & 59.7 \\
    & CFG-CD   & \textbf{70.7} \hlg{(+30.5)}   & 11.0 \hlly{(+0.6)}   & 60.4 \hlly{(+8.6)}   & 59.4 \hlr{(-1.3)}   & 24.4 \hlr{(-1.0)}   & 55.9 \hlr{(-3.8)} \\
    & \textbf{CDC (Ours)} & 44.5 \hlly{(+4.3)}   & \textbf{20.7} \hlg{(+10.3)}   & \textbf{64.6} \hlg{(+12.8)}   & \textbf{61.2} \hlly{(+0.5)}   & \textbf{33.8} \hlly{(+8.4)}   & \textbf{63.7} \hlly{(+4.0)} \\
  \cmidrule(l){2-8}
  \multirow{3}{*}{DiffuCoder 7B}
    & Vanilla      & 43.3   & 24.4   & 56.1   & 53.9   & 28.2   & 53.1 \\
    & CFG-CD   & 49.4 \hlly{(+6.1)}   & 25.0 \hlly{(+0.6)}   & 50.0 \hlr{(-6.1)}   & 50.6 \hlr{(-3.3)}   & 27.0 \hlr{(-1.2)}   & 51.1 \hlr{(-2.0)} \\
    & \textbf{CDC (Ours)} & \textbf{50.0} \hlly{(+6.7)}   & \textbf{37.2} \hlg{(+12.8)}   & \textbf{68.3} \hlg{(+12.2)}   & \textbf{56.9} \hlly{(+3.0)}   & \textbf{41.8} \hlg{(+13.6)}   & \textbf{68.0} \hlg{(+14.9)} \\
  \cmidrule(l){2-8}
  \multirow{3}{*}{Dream-Coder 7B}
    & Vanilla      & 67.1   & 34.1   & 55.5   & 41.1   & 27.7   & 54.2 \\
    & CFG-CD   & 61.6 \hlr{(-5.5)}   & 37.8 \hlly{(+3.7)}   & 53.7 \hlr{(-1.8)}   & 37.3 \hlr{(-3.8)}   & 25.7 \hlr{(-2.0)}   & 49.1 \hlr{(-5.1)} \\
    & \textbf{CDC (Ours)} & \textbf{79.3} \hlg{(+12.2)}   & \textbf{65.2} \hlg{(+31.1)}   & \textbf{83.5} \hlg{(+28.0)}   & \textbf{72.0} \hlg{(+30.9)}   & \textbf{59.2} \hlg{(+31.5)}   & \textbf{74.8} \hlg{(+20.6)} \\
    \midrule
  \midrule

\multirow{2}{*}{DeepSeek-Coder 6.7B}
    & Vanilla      & \textbf{94.5}   & 59.8   & 80.5   & 88.2   & 59.7   & 73.0 \\
    & Reprompt & 90.9 \hlr{(-3.6)}   & \textbf{64.0} \hlly{(+4.2)}   & \textbf{82.9} \hlly{(+2.4)}   & \textbf{89.2} \hlly{(+1.0)}   & \textbf{60.7} \hlly{(+1.0)}   & \textbf{74.1} \hlly{(+1.1)} \\
  \cmidrule(l){2-8}
  \multirow{2}{*}{CodeLlama 7B}
    & Vanilla      & \textbf{90.2}   & 30.5   & 54.9   & \textbf{83.9}   & 46.1   & 61.7 \\
    & Reprompt & 81.7 \hlr{(-8.5)}   & \textbf{31.1} \hlly{(+0.6)}   & \textbf{55.5} \hlly{(+0.6)}   & 83.6 \hlr{(-0.3)}   & \textbf{47.4} \hlly{(+1.3)}   & \textbf{64.7} \hlly{(+3.0)} \\
  
  \bottomrule
  
  \end{tabular}
  }
\end{table}

This section evaluates CDC for constrained code generation under four objectives: functional correctness, syntactic validity, security, and edit localization. For functional correctness and syntactic validity, we instantiate CDC with GradGuide. For security, we instantiate CDC with MDFI. We evaluate edit localization for both instantiations to measure whether CDC concentrates corrections on constraint-relevant regions. Additional experimental details and results are provided in Appendix~\ref{app:extended_exp}. 
\vspace{-1ex}
\subsection{Code Functionality and Syntax}
\vspace{-1ex}
\label{sec:exp_func}

We first evaluate CDC on the core requirements of code generation: producing programs that compile (syntactic validity) and implement the intended behavior (functional correctness). 

\textbf{Baselines.} We evaluate CDC on three diffusion language models: Dream-Coder-7B \citep{xie2025dreamcoder}, DiffuCoder-7B \citep{gong2025diffucoder}, and Dream-7B \citep{ye2025dream}. We compare against two diffusion baselines: Vanilla, which uses the base denoising sampler without constraint intervention, and CFG-CD \citep{muendler2025cfgdiffusion}, which uses grammar-constrained diffusion decoding for syntax-level constraints. We also compare against two autoregressive code models, DeepSeek-Coder-Instruct-6.7B \citep{guo2024deepseekcoder} and CodeLlama-7B \citep{roziere2024codellamaopenfoundation}, under vanilla and reprompt configurations. Reprompt first performs vanilla generation; if the output fails the constraint, test pass/fail feedback is used to launch one additional generation round. \textbf{Benchmarks.} We evaluate C++ program synthesis on HumanEval-X C++ (164 tasks) and MBPP-C++ (397 tasks) \citep{chen2021codex,austin2021program,zheng2023codegeex}. 
\textbf{Evaluation Metrics.}
Functionality is measured by pass@1 and pass@10: a task is solved if at least one of 1 or 10 generated samples compiles and passes all tests. Syntax is the compile-success rate.

\begin{table*}[t]
    \caption{Security-aware code generation on CWEval and LLMSecEval+. Parenthesized values give the absolute change relative to the vanilla model: \hlg{green} marks gains of $\geq$10\%, \hlly{yellow} gains of $<$10\%, and \hlr{red}  regressions. Within each model, the best performance is \textbf{bolded}.}
    \label{tab:main_security}
    \centering

    \setlength{\tabcolsep}{3pt}
    \renewcommand{\arraystretch}{0.95}
    \scalebox{0.69}{
    \begin{tabular}{@{}l l lllll lllll@{}}
    \toprule
    & & \multicolumn{5}{c}{\textbf{CWEval}} & \multicolumn{5}{c}{\textbf{LLMSecEval+}} \\
    \cmidrule(lr){3-7}\cmidrule(lr){8-12}
    \textbf{Model} & \textbf{Method}
        & func@1 & sec@1 & fs@1 & fs@5 & fs@10
        & func@1 & sec@1 & fs@1 & fs@5 & fs@10 \\
    \midrule
    \multirow{3}{*}{Diffusion}
        & Vanilla
            & 26.9 & 18.5 & 12.0 & 25.9 & 26.9
            & 32.0 & 54.7 & 14.7 & 22.0 & 24.0 \\
        & Sec prompt
            & 26.9 \hlly{(+0)} & 19.4 \hlly{(+1)} & 13.9 \hlly{(+2)} & 23.2 \hlr{(-3)} & 27.8 \hlly{(+1)}
            & \textbf{34.7} \hlly{(+3)} & 58.7 \hlly{(+4)} & 16.0 \hlly{(+1)} & 22.7 \hlly{(+1)} & 26.7 \hlly{(+3)} \\
        & \textbf{CDC}
            & \textbf{39.8} \hlg{(+13)} & \textbf{41.7} \hlg{(+23)} & \textbf{34.3} \hlg{(+22)} & \textbf{49.1} \hlg{(+23)} & \textbf{51.9} \hlg{(+25)}
            & 30.7 \hlr{(-1)} & \textbf{80.7} \hlg{(+26)} & \textbf{24.7} \hlg{(+10)} & \textbf{40.7} \hlg{(+19)} & \textbf{45.3} \hlg{(+21)} \\
    \midrule
    \midrule
    \multirow{2}{*}{AR}
        & Vanilla
            & \textbf{48.2} & 24.1 & 21.3 & 36.1 & 40.7
            & 37.3 & 55.3 & 21.3 & 30.7 & 38.0 \\
        & Reprompt
            & 44.4 \hlr{(-4)} & \textbf{47.2} \hlg{(+23)} & \textbf{39.8} \hlg{(+19)} & \textbf{63.0} \hlg{(+27)} & \textbf{68.5} \hlg{(+28)}
            & \textbf{40.0} \hlly{(+3)} & \textbf{64.7} \hlly{(+9)} & \textbf{26.0} \hlly{(+5)} & \textbf{40.0} \hlly{(+9)} & \textbf{48.7} \hlg{(+11)} \\
    \bottomrule
    \end{tabular}
    }
\end{table*}

Table~\ref{tab:main_functional} reports syntax and functional correctness on HumanEval-X C++ and MBPP-C++. Across diffusion backbones, CDC improves both pass@1 and pass@10 over vanilla diffusion, with the
largest gains on stronger code-diffusion models. On HumanEval-X C++,
CDC improves Dream-Coder 7B from 34.1\% to 65.2\%
pass@1, while also increasing compile success from 67.1\%
to 79.3\%. On MBPP-C++, the same model improves from 27.7\% to
59.2\% pass@1, with compile success increasing from
41.1\% to 72.0\%. CDC also improves weaker diffusion
backbones, increasing Dream 7B from 10.4\% to 20.7\% on HumanEval-X
and DiffuCoder 7B from 24.4\% to 37.2\%. The improvements are consistent at pass@10. 

While CFG-CD improves grammar or compile
validity in several cases, it does not consistently improve functional correctness. For example, on MBPP-C++, CFG-CD decreases Dream-Coder 7B pass@1 from 27.7\% to
25.7\%, whereas CDC increases it to 59.2\%. This difference
suggests that syntax constraints alone are insufficient for program synthesis:
CDC improves both validity and test-pass rates because its surrogate
signal is trained from execution outcomes rather than grammar membership
alone.

Compared with autoregressive baselines, CDC substantially narrows the functionality gap while retaining the advantages of diffusion-style correction. On HumanEval-X C++, CDC with Dream-Coder 7B achieves 65.2\% pass@1 and 83.5\% pass@10, outperforming all 7B autoregressive baselines, including DeepSeek-Coder with re-prompting at 64.0\% pass@1 and 82.9\% pass@10. On MBPP-C++, CDC reaches 59.2\% pass@1 and 74.8\% pass@10, approaching DeepSeek-Coder and substantially outperforming CodeLlama. These results suggest that denoising-time constraint integration can make diffusion code models competitive with strong autoregressive code generators, without relying on full-program re-generation after feedback. Additional regression analysis and ablations are provided in Appendix~\ref{app:additional_results}.

\vspace{-1ex}
\subsection{Code Generation Security}
\vspace{-1ex}
\label{sec:exp_security}

We next evaluate whether CDC can improve security-oriented code generation, where outputs are evaluated to satisfy both security and functionality constraints. This setting is challenging as security fixes often require localized semantic changes while preserving the surrounding program behavior.

\textbf{Baselines.} We use Dream-Coder-7B as the diffusion backbone for CDC and DeepSeek-Coder-Instruct-6.7B as the autoregressive baseline. For autoregressive models, we compare vanilla generation with AR re-prompting \citep{chen2023teachinglargelanguagemodels}, where an insecure or failing solution is returned to the model with feedback from a static program analyzer and the model regenerates a complete program. For diffusion models, we compare against vanilla diffusion and security prompting.
\textbf{Benchmarks.} We evaluate CDC on CWEval and LLMSecEval+ \citep{peng2025cweval,tony2023llmseceval} (see Appendix~\ref{sec:benchmarks}). \textbf{Metrics.} We report func@1 and sec@1, the fractions of single samples that pass the functional and security oracles, respectively, and joint func-sec@$k$ (fs@$k$) for $k\in\{1,5,10\}$, where a task is solved if any of $k$ sampled outputs is both functional and secure.

Table~\ref{tab:main_security} shows that CDC substantially improves security-aware generation over vanilla diffusion and secure prompting. On CWEval, CDC increases fs@1 from 12.0\% to 34.3\%, nearly tripling the joint functionality-security success rate, and lifts both component metrics in tandem (func@1 26.9\%$\to$39.8\%, sec@1 18.5\%$\to$41.7\%). On LLMSecEval+, CDC improves security most strongly, increasing sec@1 from 54.7\% for vanilla diffusion and 58.7\% with secure prompting to 80.7\%---higher than every other method in the table, including AR re-prompting (64.7\%)---while also improving fs@1 from 14.7\% to 24.7\%. The improvement extends to multi-sample regimes: CDC raises CWEval fs@5/fs@10 from 25.9\%/26.9\% to 49.1\%/51.9\%, and matches AR re-prompting on LLMSecEval+ fs@5 (40.7\% vs.\ 40.0\%).

Compared with AR re-prompting, CDC is competitive despite using a weaker diffusion backbone for functionality. AR re-prompting obtains the highest overall fs@1 on both benchmarks, but CDC closes much of the gap while performing interventions during generation rather than regenerating complete programs after failure. The per-language CWEval results (Table~\ref{tab:per_language_security_table}, Appendix~\ref{app:additional_results}) show that CDC is especially strong on Python, where it reaches 72.0\% fs@1, and remains competitive on JavaScript, C++, and Go, suggesting that localized mid-diffusion repair can improve security without relying only on post-hoc complete-program correction. Additional efficiency analysis, regression analysis and ablations are provided in Appendix~\ref{app:additional_results}.

\vspace{-1ex}
\subsection{Edit Locality}
\vspace{-1ex}
\label{sec:exp_locality}

\begin{figure}[t]
    \centering
    \includegraphics[width=0.80\linewidth]{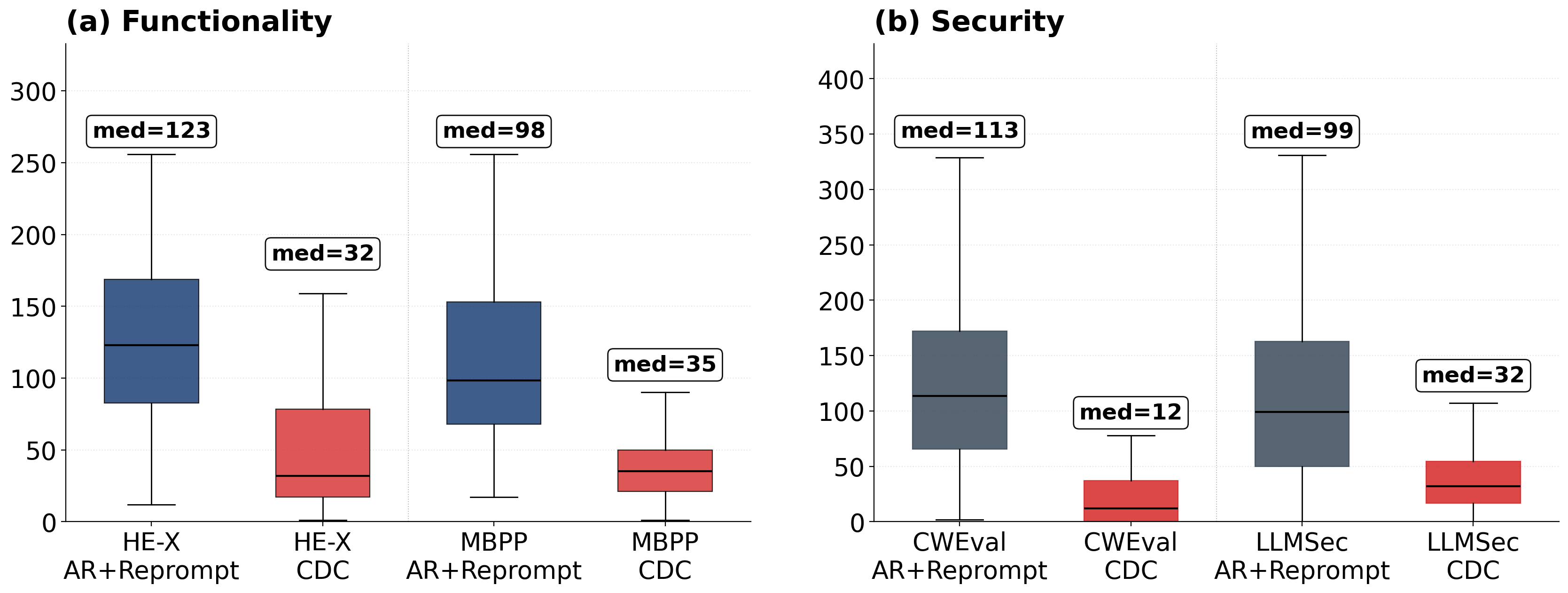}
    \caption{Edited tokens per correction attempt (fewer means more localized edits): (a) functionality corrections on HumanEval-X and MBPP, and (b) security corrections on CWEval and LLMSecEval+.}
    \vspace{-15pt}
    \label{fig:main_local}
\end{figure}

In addition to task success, we evaluate whether CDC performs targeted constraint correction rather than broad regeneration. We measure edit locality across the same functional-correctness and security benchmarks described above. For each intervention, we compare the constrained output to the corresponding unconstrained/base output and report edit locality statistics, including the fraction of changed tokens, number of changed spans or edit clusters, and tokens inserted or remasked. We also measure corrective cost, including the number of additional denoising interventions, remasked tokens, and generated tokens. These metrics are used to assess whether CDC improves constraint satisfaction while concentrating changes on constraint-relevant regions and avoiding repeated full-program regeneration. 

Figure~\ref{fig:main_local} reports the number of edited tokens for security and functionality corrections, where fewer edited tokens indicate a more targeted intervention. For security, CDC is substantially more local than AR re-prompting on both benchmarks: AR re-prompting rewrites a median of 113 tokens on CWEval and 99 tokens on LLMSecEval+, while CDC edits only 12 and 32 tokens, respectively. This supports the main motivation of CDC: because diffusion states remain editable during generation, analyzer feedback can be applied directly to the vulnerable region without regenerating the full program.

For functionality, CDC also produces more localized corrections than AR re-prompting. On HumanEval-X C++, CDC edits a median of 32 tokens compared with 123 tokens for AR re-prompting. On MBPP-C++, CDC edits a median of 35 tokens compared with 98 tokens for AR re-prompting. Thus, across both security and functionality settings, CDC reduces the amount of code rewritten during correction, suggesting that its gains come from focused denoising-time interventions rather than full-program repair. Additional locality analysis and ablations are included in Appendix~\ref{app:additional_results}.

\vspace{-1ex}
\subsection{Inference Cost and Surrogate Reliability}
\vspace{-1ex}
\label{sec:exp_cost}

\begin{table}[t]
\centering
\def\ratio#1{{\scriptsize\textcolor{gray!75!black}{(#1$\times$)}}}%
\caption{Inference cost of CDC versus vanilla decoding (Van.) under identical settings. Gray values give the ratio to vanilla; memory is the increase in peak GPU memory.}
\label{tab:inference_cost}
\footnotesize
\setlength{\tabcolsep}{4.5pt}
\renewcommand{\arraystretch}{1.08}
\begin{tabular}{@{}l rl rl r rl@{}}
\toprule
& \multicolumn{2}{c}{\textbf{Mean latency (s)}} & \multicolumn{2}{c}{\textbf{Median latency (s)}} & \textbf{Peak mem.} & \multicolumn{2}{c}{\textbf{Success (\%)}} \\
\cmidrule(lr){2-3} \cmidrule(lr){4-5} \cmidrule(lr){7-8}
\textbf{Setting} & Van. & CDC & Van. & CDC & $\Delta$ & Van. & CDC \\
\midrule
\multicolumn{8}{@{}l}{\textit{GradGuide on HumanEval-X C++ (pass@1)}} \\
Dream-Coder 7B & 4.23 & 7.99 \ratio{1.89} & 4.26 & 6.10 \ratio{1.43} & +53.0\% & 34.1 & \textbf{65.2} \hlg{(+31.1)} \\
Dream 7B & 4.18 & 7.20 \ratio{1.72} & 4.21 & 5.93 \ratio{1.41} & +57.8\% & 10.4 & \textbf{20.7} \hlg{(+10.3)} \\
DiffuCoder 7B & 4.16 & 8.37 \ratio{2.01} & 4.19 & 6.16 \ratio{1.47} & +60.5\% & 24.4 & \textbf{37.2} \hlg{(+12.8)} \\
\midrule
\multicolumn{8}{@{}l}{\textit{MDFI with Dream-Coder 7B (fs@1)}} \\
CWEval & 12.98 & 13.28 \ratio{1.02} & 12.85 & 12.97 \ratio{1.01} & +4.2\% & 12.0 & \textbf{34.3} \hlg{(+22.3)} \\
LLMSecEval+ & 10.12 & 10.73 \ratio{1.06} & 9.62 & 11.00 \ratio{1.14} & +2.6\% & 14.7 & \textbf{24.7} \hlg{(+10.0)} \\
\bottomrule
\end{tabular}
\vspace{-6pt}
\end{table}

We also evaluate the end-to-end inference cost of CDC. Table~\ref{tab:inference_cost} reports per-sample latency and peak GPU memory, with each method using the same backbone, decoding configuration, hardware, and 256-step schedule as its vanilla baseline (Appendix~\ref{app:inference_cost}). GradGuide intervenes on 66--84\% of samples at 48--53\,ms per surrogate pass, raising mean latency to 1.72--2.01$\times$ that of vanilla decoding and peak memory by 53--61\%; the median latency rises only to 1.41--1.47$\times$, as the cost concentrates in trajectories that require repeated correction. In return, GradGuide improves pass@1 by 31.1 and 31.5 points on HumanEval-X C++ and MBPP-C++, compared with 3.1 and 3.5 points for Best-of-$N$ sampling ($N\!=\!2$) and 4.2 and 1.0 points for AR re-prompting. MDFI adds only 2.3--6.1\% latency from seven CPU-based analyzer calls (46\,ms in total) and achieves 2.79$\times$ and 1.58$\times$ the successful-output throughput of vanilla decoding on CWEval and LLMSecEval+.

Given the additional inference cost of GradGuide, we next examine the reliability of the surrogate that guides its corrections. The surrogate is trained once on CodeContests problems disjoint from both benchmarks, with a reported training cost of 1.8 GPU-hours. It accurately predicts execution outcomes on HumanEval-X C++ and MBPP-C++ (Spearman $\rho$ of 0.848 and 0.837, ECE of 0.024 and 0.030), whereas substituting the non-fine-tuned backbone ($\rho$ of 0.056 and 0.117) lowers pass@1 by 8.5 and 5.3 points relative to vanilla decoding. These results highlight the importance of surrogate quality in achieving effective guidance (Appendix~\ref{app:surrogate_eval}).

\vspace{-1ex}
\section{Limitations}
\label{sec:limitations}
\vspace{-1ex}
CDC depends on the quality and coverage of the feedback mechanisms used during denoising. Specifically, GradGuide relies on a learned surrogate, which may provide noisy or misleading estimates of functional correctness, while MDFI relies on static analysis, which may miss vulnerabilities or produce false positives. The empirical evaluation is also limited to a fixed set of benchmarks, languages, and diffusion backbones, so performance may differ for larger codebases, longer-context tasks, or more complex software-engineering workflows. Finally, although CDC performs more localized edits than post-hoc regeneration, it still adds inference-time overhead from surrogate evaluation, static analysis, and constraint-aware correction.

\vspace{-1ex}
\section{Conclusion}
\vspace{-1ex}
This paper introduced CDC, an inference-time neurosymbolic framework that integrates program-level constraints into the reverse denoising process of diffusion language models without retraining the base model. CDC uses the global, editable states exposed during sampling to evaluate intermediate programs, localize constraint feedback, and apply targeted corrections that reduce violation while staying close to the base denoiser. The framework supports both soft and hard feedback: GradGuide uses soft surrogate signals for functional correctness, while MDFI uses hard symbolic program-analysis signals for security repair through remasking, insertion, and feedback injection. Across functional and security-oriented benchmarks, CDC improves test pass rates, compile success, and joint functionality-security success over standard diffusion decoding and syntax-only constrained diffusion. These results show that discrete diffusion language models are a natural substrate for constrained code generation, enabling focused correction during generation instead of post-hoc filtering or whole-program regeneration.

\section*{Broader Impact}
\label{sec:broader_impact}
This work aims to improve the reliability and security of code generation by steering diffusion models toward programs that better satisfy functional and security constraints. Potential benefits include fewer syntax errors, higher task correctness, and reduced vulnerabilities in generated code. However, stronger code generation is dual-use and could be misused if constraint mechanisms are adapted toward harmful objectives.

\clearpage
\begin{ack}
This research is partially supported by NSF awards 2533631, 2401285, 2334936, and 2226816, and by DARPA under Contract No. $\#$HR0011252E005. The authors acknowledge the Research Computing at the University of Virginia. Any opinions, findings, conclusions, or recommendations expressed in this material are those of the authors and do not necessarily reflect the views of NSF or DARPA. This material is based upon work supported by the National Science Foundation Graduate Research Fellowship Program under Grant No.~2234693. Any opinions, findings, and conclusions or recommendations expressed in this material are those of the authors and do not necessarily reflect the views of the National Science Foundation.

\medskip
\noindent\textbf{Competing Interests.}
The authors declare no competing financial interests relevant to this work. All research funding and computational support are disclosed above.
\end{ack}

\bibliographystyle{unsrtnat}
\bibliography{references}

\appendix
\section{Related work (Extended)}
With the success of diffusion models in continuous domains, especially in image and video generation~\cite{ho2020denoisingdiffusionprobabilisticmodels, rombach2022highresolutionimagesynthesislatent}, recent work has extended diffusion principles to discrete data. Early work by Austin et al.~\cite{austin2023structureddenoisingdiffusionmodels} introduced discrete denoising diffusion models, which generalize multinomial diffusion with structured transition operators for discrete state spaces. Sahoo et al.~\cite{sahoo2024simpleeffectivemaskeddiffusion} subsequently showed that masked discrete diffusion with an absorbing-state corruption process can be effective for language modeling, deriving a simplified Rao-Blackwellized training objective and establishing a strong masked diffusion baseline that approaches autoregressive perplexity. Concurrently, Shi et al.~\cite{NEURIPS2024_bad233b9} introduced MD4, which adopts the same masked diffusion formulation and objective and achieves comparable state-of-the-art language-modeling results. More recently, these models have been scaled to large language model regimes. Ye et al.~\cite{ye2025dream} introduced Dream 7B, a scalable diffusion large language model that performs competitively with similarly sized autoregressive models while emphasizing parallel refinement, planning, and flexible inference. In the coding domain, Gong et al.~\cite{gong2025diffucoder} developed DiffuCoder, a 7B diffusion language model specialized for code generation, and showed that diffusion-specific decoding analysis and reinforcement-learning-based post-training can further improve coding performance. More recently, Xie et al.~\cite{xie2025dreamcoder} introduced Dream-Coder 7B, an open diffusion language model for code generation with emergent any-order generation capabilities, demonstrating that diffusion language models can adapt their decoding behavior across coding tasks while achieving competitive performance on standard code benchmarks.

Controllable generation is a central challenge in generative AI, particularly for code generation, where outputs are expected not only to be plausible, but also to satisfy validity, correctness, safety, and other program-level properties. In autoregressive language models, this has been studied through constrained decoding, reranking, rejection sampling, and iterative self-correction~\cite{hu-etal-2019-improved, ni2023leverlearningverifylanguagetocode}. These methods typically enforce constraints by restricting next-token decisions, selecting among multiple completed candidates, or repairing an initial output using verifier or execution feedback. Specific to code generation, this line includes self-debugging and test-driven refinement~\cite{chen2023teachinglargelanguagemodels}, type-constrained decoding~\cite{M_ndler_2025}, and security-aware code evaluation and repair~\cite{bhatt2024cyberseceval2widerangingcybersecurity}. More broadly, these approaches inherit the structure of autoregressive decoding: early decisions propagate forward, revising a constraint-relevant region often requires regenerating substantial suffixes, and program-level constraints are therefore typically handled through token-local controls or post-hoc repair.

In contrast, constraint integration is particularly suitable for discrete diffusion models, as they refine the entire sequence over multiple denoising steps, creating natural opportunities for sequence-level constraint enforcement during sampling. This has led to several works on controllable generation. Schiff et al.~\cite{schiff2025simpleguidancemechanismsdiscrete} derive classifier-based and classifier-free guidance for discrete diffusion models, adapting standard diffusion guidance mechanisms to discrete state spaces. For protein design, Gruver et al.~\cite{gruver2023proteindesignguideddiscrete} introduce diffusioN Optimized Sampling (NOS), which performs guidance through gradients in the denoiser hidden states and supports constrained sequence optimization. Other techniques~\cite{cardei2025constraineddiscretediffusion, christopher2025neurosymbolicgenerativediffusionmodels} formulate diffusion sampling as constrained optimization and apply projection-based updates during reverse denoising. 

Specific to code generation, M\"undler et al.~\cite{muendler2025cfgdiffusion} present a constrained decoding method for diffusion LLMs under context-free grammars, showing that syntax-level constraints for programming languages can be enforced during out-of-order denoising with improved syntactic correctness. Complementary to such decoding-time constraints, TreeDiff~\cite{zeng-etal-2026-treediff} injects structure at training time, masking the tokens of key abstract-syntax-tree nodes rather than random tokens so that the model learns to respect syntactic boundaries.
The work presented in this paper is most closely related to this line of research, but targets a harder code-generation setting in which constraints are semantic and security-related. In this setting, prior guidance-based and syntax-level constrained decoding methods are not sufficient, as program semantic correctness and security generally cannot be captured by local token preferences or grammar constraints alone.

Table~\ref{tab:related_work_summary} summarizes the closest lines of work and the dimensions that most directly separate CDC from prior approaches.

\begin{table}[t]
\centering
\caption{Comparison of CDC with related approaches.}
\label{tab:related_work_summary}
\footnotesize
\setlength{\tabcolsep}{3pt}
\renewcommand{\arraystretch}{0.95}
\begin{tabular}{@{}p{4.55cm} c p{1.15cm} p{2.15cm} c@{}}
\toprule
\textbf{Approach} & \textbf{Base} & \textbf{When} & \textbf{Constraint Scope} & \makecell{\textbf{Semantic /}\\\textbf{Verifier}} \\
\midrule
\makecell[l]{AR constrained decoding,\\reranking, self-correction~\\\cite{hu-etal-2019-improved,ni2023leverlearningverifylanguagetocode,chen2023teachinglargelanguagemodels,bhatt2024cyberseceval2widerangingcybersecurity}} & AR & \makecell[c]{token /\\post} & \makecell[l]{tests, execution,\\repair feedback} & \makecell[c]{post-hoc\\only} \\
\makecell[l]{Type-constrained\\decoding~\\\cite{M_ndler_2025}} & AR & token & \makecell[l]{syntax,\\types} & No \\
\makecell[l]{Discrete diffusion\\guidance~\\\cite{schiff2025simpleguidancemechanismsdiscrete,gruver2023proteindesignguideddiscrete}} & Diff. & denoise & \makecell[l]{soft or gradient-\\based objectives} & Limited \\
\makecell[l]{Optimization-based\\constrained diffusion~\\\cite{cardei2025constraineddiscretediffusion,christopher2025neurosymbolicgenerativediffusionmodels}} & Diff. & denoise & \makecell[l]{projection /\\optimization} & Limited \\
\makecell[l]{CFG-constrained diffusion\\for code~\\\cite{muendler2025cfgdiffusion}} & Diff. & denoise & \makecell[l]{syntax\\(CFG)} & No \\
\rowcolor{blue!6}
\textbf{CDC (ours)} & Diff. & denoise & \makecell[l]{syntax, semantics,\\verifier feedback} & \textbf{Yes} \\
\bottomrule
\end{tabular}
\end{table}

\section{Extended Experimental Details and Results}
\label{app:extended_exp}

\subsection{Benchmarks}
\label{sec:benchmarks}

\paragraph{Syntax and functionality.} C++ program synthesis is evaluated with HumanEval-X C++ and MBPP-C++
\citep{chen2021codex,austin2021program,zheng2023codegeex}.
HumanEval-X extends HumanEval to multiple programming languages with
hand-written reference solutions and execution tests; we use its 164 C++
tasks. MBPP-C++ contains 397 C++ translations of MBPP-style programming
problems. For each problem, the model receives the natural-language
description and function signature, and must generate a complete solution.

We report compile success (\(\mathrm{Syntax}\)), functional
pass@1, and pass@10. A sample is functionally
correct if it compiles and passes all unit tests. For pass@1, every model is run deterministically at temperature 0 with seed 0, 256 denoising / decoding steps, and a maximum of 512 generated tokens. For pass@10 we draw 10 independent samples (seeds 0--9, otherwise the same step and length budget) and report the unbiased estimator $\mathrm{pass@}k=\mathbb{E}\bigl[1-\binom{n-c}{k}/\binom{n}{k}\bigr]$, with $n\!=\!10$ and $c$ the number of correct samples per task.

\paragraph{Security.} Security-oriented code generation is evaluated on CWEval and LLMSecEval
\citep{peng2025cweval,tony2023llmseceval}. CWEval additionally evaluates functionality and security on the same programming tasks, making it suitable for testing a method's impact on correctness under security objectives. This dataset contains 108 prompts spanning C, C++, Go, JavaScript, and Python. LLMSecEval contains natural-language prompts associated with common CWE patterns; LLMSecEval+ is constructed by adding executable functional and security oracles to 150 LLMSecEval prompts.

For security, we report three pass@1 metrics. Let \(F_i\) indicate that
sample \(i\) passes the functional oracle, and let \(S_i\) indicate that it
passes the security oracle. Then
\[
    \mathrm{func@1}=\mathbb{E}[F_i],
    \qquad
    \mathrm{sec@1}=\mathbb{E}[S_i],
    \qquad
    \mathrm{func\mbox{-}sec@1}=\mathbb{E}[F_iS_i].
\]
The joint metric func-sec@1 is our primary security
metric, since generated code must be both secure and functionally correct.

\subsection{Implementation Details}
\label{sec:implementation-details}

\textbf{GradGuide.} GradGuide uses a Qwen2.5-Coder-1.5B surrogate \citep{hui2024qwen25coder} trained
on CodeContests-derived programs \citep{li2022alphacode} with execution
labels (Appendix~\ref{app:surrogate_eval}). The surrogate predicts a continuous correctness score from soft token
embeddings and is used for both localization and ALM correction. We apply GradGuide adaptively during reverse denoising, using adaptive editing to localize and refine constraint-relevant regions. Specifically, surrogate-guided proposal corrections are activated when needed, while high-saliency token positions are expanded into local syntactic neighborhoods and remasked when the constraint-violation and edit-budget conditions are satisfied.

\textbf{MDFI.} MDFI invokes the static-analysis evaluator at fixed denoising checkpoints.
The analyzer constructs a tolerant partial Code Property Graph
\citep{yamaguchi2014modeling} using Tree-sitter-style incremental parsing
\citep{treesitter}, extracts dataflow witnesses with bounded source-to-sink
search, obtains structural risk with the product FSM when exact labels are
masked, selects semantic neighborhoods from the AST under a token budget,
and injects compact remediation feedback into a preallocated prompt buffer.

\textbf{Efficiency measurements.} We measure both outcome quality and corrective cost. For GradGuide, we
report total tokens generated by the model, tokens per instance, tokens per
passing solution, tokens rewritten per failed instance, and fraction of the
program body edited. For MDFI, we additionally measure edit span locality and
number of edit clusters. AR re-prompting is counted as regenerating the full
program body, while CDC counts only the regions that are remasked or inserted
during the denoising trajectory. Efficiency measurements and per-prompt analyses in Appendix~\ref{app:additional_results} each use one generation per prompt, independently sampled across analyses from the multi-sample runs underlying pass@10 in Table~\ref{tab:main_functional} and fs@10 in Table~\ref{tab:main_security} for all methods.

\subsection{Additional Experimental Results}
\label{app:additional_results}

\label{sec:experimental-results}
\begin{figure*}[t]
    \centering
    \includegraphics[width=\textwidth]{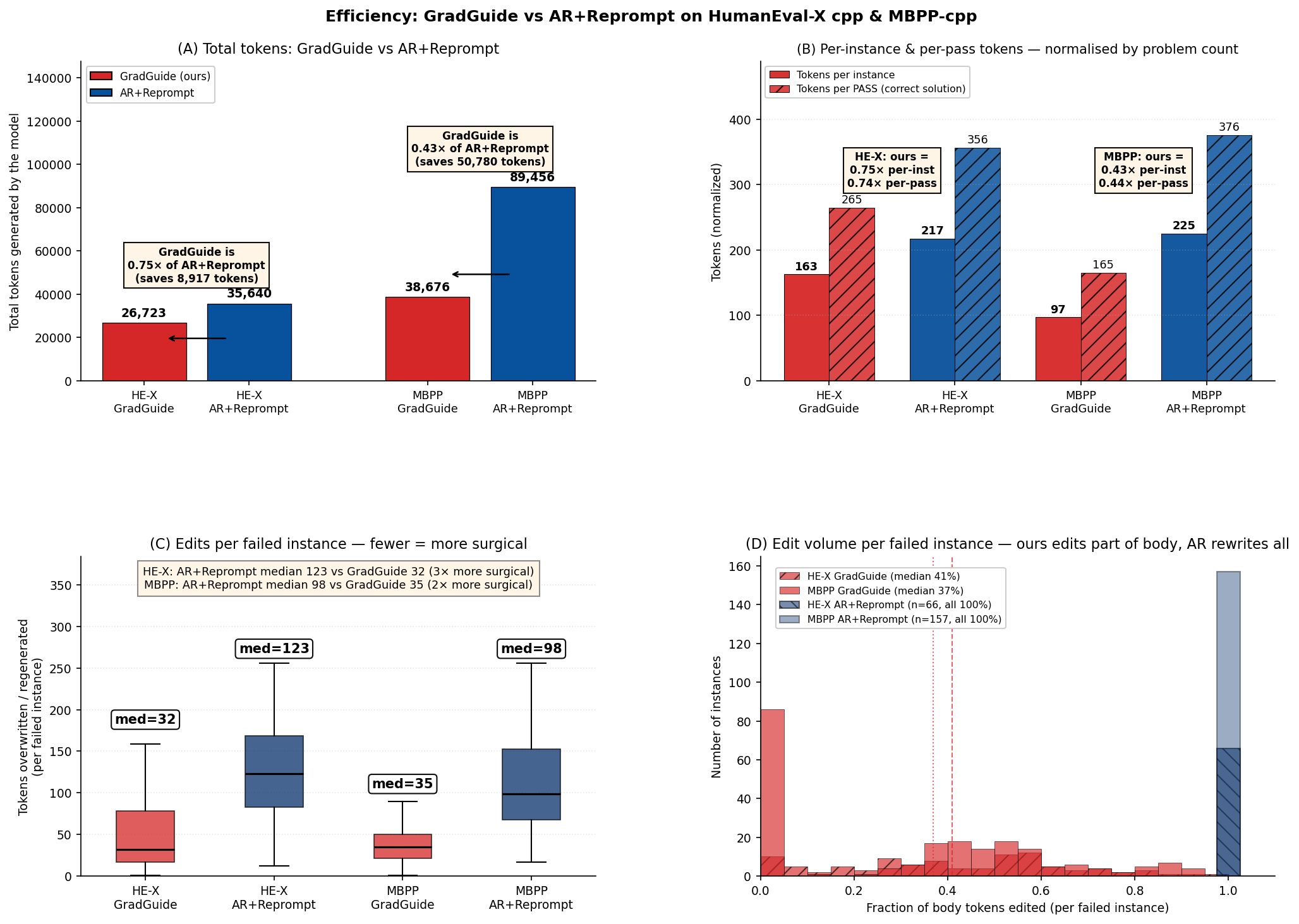}
    \caption{
    Efficiency and locality of CDC vs.\ AR+Reprompt
    }
    \label{fig:gardguide-efficiency}
\end{figure*}

\begin{table}[t]
    \caption{Per-language \textbf{fs@1} on CWEval (108 prompts total). Parenthesized values give the absolute change relative to the corresponding model's Vanilla row: \hlly{yellow} marks an improvement of $<$10 points, \hlg{green} marks an improvement of $\geq$10 points, and \hlr{red} marks a regression. Within each model, the strongest value per language is \textbf{bolded}.}
    \label{tab:per_language_security_table}
    \centering
    \scriptsize
    \setlength{\tabcolsep}{4pt}
    \renewcommand{\arraystretch}{0.95}
    \scalebox{0.95}{
    \begin{tabular}{@{}l l llllll@{}}
    \toprule
    \textbf{Model} & \textbf{Method} & py (25) & js (23) & c (20) & cpp (21) & go (19) & ALL (108) \\
    \midrule
    \multirow{3}{*}{Diffusion}
        & Vanilla         & 24.0 & 13.0 & 0.0 & 14.3 & 5.3 & 12.0 \\
        & + sec prompt    & 28.0 \hlly{(+4)} & 17.4 \hlly{(+4)} & 0.0 \hlly{(+0)} & 14.3 \hlly{(+0)} & 5.3 \hlly{(+0)} & 13.9 \hlly{(+2)} \\
        & \textbf{MDFI (Ours)} & \textbf{72.0} \hlg{(+48)} & \textbf{34.8} \hlg{(+22)} & \textbf{5.0} \hlly{(+5)} & \textbf{28.6} \hlg{(+14)} & \textbf{21.1} \hlg{(+16)} & \textbf{34.3} \hlg{(+22)} \\
    \midrule
    \midrule
    \multirow{2}{*}{AR}
        & Vanilla  & 44.0 & 21.7 & 10.0 & 14.3 & 10.5 & 21.3 \\
        & Reprompt & \textbf{64.0} \hlg{(+20)} & \textbf{43.5} \hlg{(+22)} & \textbf{25.0} \hlg{(+15)} & \textbf{33.3} \hlg{(+19)} & \textbf{26.3} \hlg{(+16)} & \textbf{39.8} \hlg{(+19)} \\
    \bottomrule
    \end{tabular}
    }
\end{table}

In this section we present additional detailed results.

\begin{figure*}[t]
    \centering
    \includegraphics[width=\textwidth]{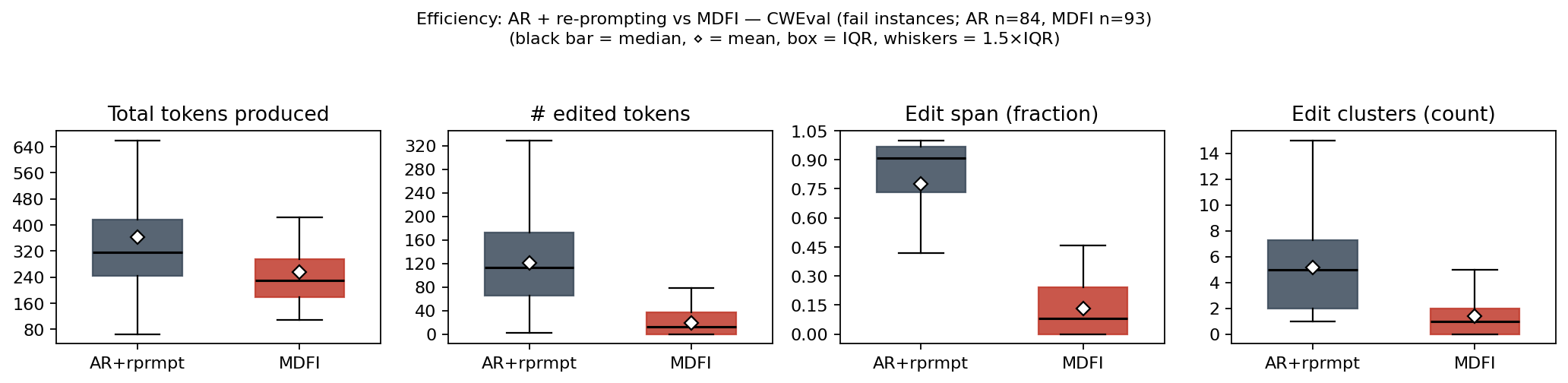}
    \caption{
    CWEval efficiency comparison between AR re-prompting and MDFI. MDFI lowers
    pipeline token cost, edited-token count, edit span, and edit clusters by
    repairing localized vulnerable regions rather than regenerating the entire
    program.
    }
    \label{fig:mdfi-efficiency-box}
\end{figure*}

\begin{figure*}[t]
    \centering
    \includegraphics[width=\textwidth]{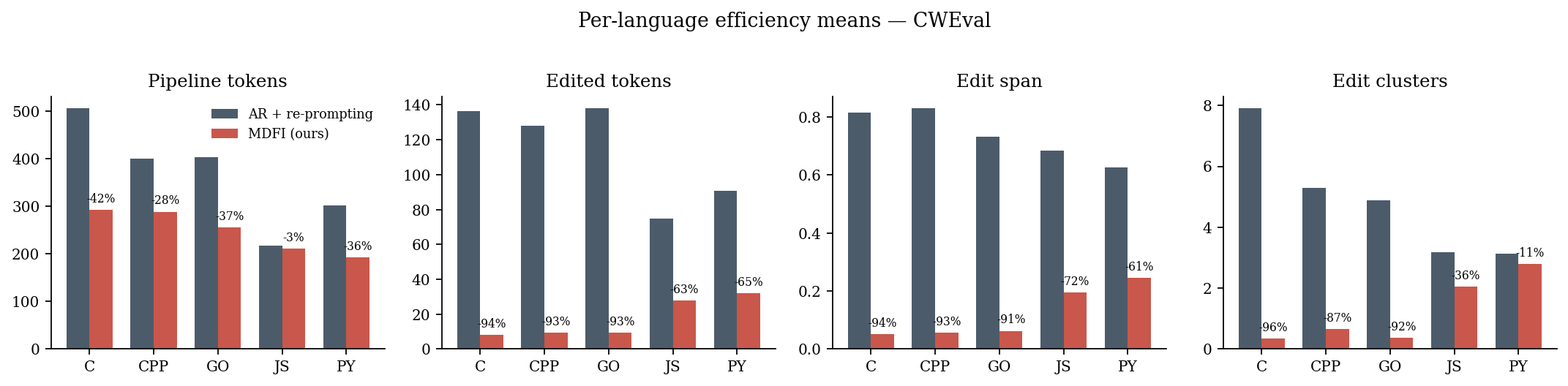}
    \caption{
    Per-language efficiency means on CWEval. MDFI substantially reduces edited
    tokens, edit span, and edit clusters across C, C++, Go, JavaScript, and
    Python.
    }
    \label{fig:mdfi-efficiency-lang}
\end{figure*}

\subsubsection{CDC is substantially more local and token-efficient than AR re-prompting}
\label{sec:gradguide-efficiency}

Figure~\ref{fig:gardguide-efficiency} compares GradGuide to AR plus one
re-prompt on HumanEval-X C++ and MBPP-C++. On HumanEval-X, AR plus one
re-prompt generates 35{,}640 total model tokens, whereas GradGuide generates
26{,}723 tokens, a 0.75$\times$ cost ratio and a savings of 8{,}917 tokens.
On MBPP-C++, AR plus one re-prompt generates 89{,}456 tokens, whereas
GradGuide generates 38{,}676 tokens, a 0.43$\times$ cost ratio and a
savings of 50{,}780 tokens.

Normalizing by problem count and by successful solution gives the same
conclusion. On HumanEval-X, GradGuide uses 163 tokens per instance and
265 tokens per passing solution, compared with 217 and 356 for AR
re-prompting. On MBPP-C++, GradGuide uses 97 tokens per instance and
165 tokens per passing solution, compared with 225 and 376 for AR
re-prompting. Thus the efficiency gain is not merely due to differences in
benchmark size; it persists when normalized by both instances and successful
outputs.

The locality analysis explains the efficiency gain. For failed instances that
require correction, AR re-prompting corrects a median of 123 tokens on
HumanEval-X and 98 tokens on MBPP-C++, while GradGuide edits medians of
32 and 35 tokens, respectively. In other words, GradGuide is about 3$\times$
more surgical on HumanEval-X and 2$\times$ more surgical on MBPP-C++. At
the body-fraction level, AR re-prompting rewrites essentially the entire
program, whereas GradGuide concentrates edits on localized regions identified
by surrogate saliency. This supports the focused-editing hypothesis: CDC
achieves its gains by revising the parts of the denoising trajectory most
relevant to the constraint, not by repeatedly sampling complete programs.

\label{sec:mdfi-efficiency}

Figure~\ref{fig:mdfi-efficiency-box} compares MDFI with AR re-prompting on
CWEval. AR re-prompting produces roughly a full second program, with a mean
pipeline cost around 3.1$\times$10\textsuperscript{2} tokens. MDFI reduces the mean pipeline
cost to roughly 2.1$\times$10\textsuperscript{2} tokens. The difference is larger for edited
tokens: AR re-prompting edits on the order of an entire solution, while MDFI
edits a small localized span. Median edit span drops from 0.89 of the program
body under AR re-prompting to below 0.1 under MDFI, and median edit
clusters drop from 4 to 1.

The per-language analysis in Figure~\ref{fig:mdfi-efficiency-lang} confirms
that locality is robust across languages. MDFI reduces edited tokens by 94\%
for C, 93\% for C++, 93\% for Go, 63\% for JavaScript, and 65\% for Python.
Edit-span reductions are similarly large: 94\% for C, 93\% for C++, 91\% for
Go, 72\% for JavaScript, and 61\% for Python. Edit-cluster reductions are
strongest in C/C++/Go, where vulnerable patterns tend to be localized around
library calls or pointer/string operations; they are smaller in Python and
JavaScript, where secure fixes often require changing a higher-level API
pattern across multiple nearby tokens. Even in those languages, MDFI remains
substantially more local than whole-program re-prompting.

\subsubsection{Inference cost and throughput}
\label{app:inference_cost}

Tables~\ref{tab:app_gradguide_cost}--\ref{tab:app_mdfi_cost} give the full measurements summarized in Section~\ref{sec:exp_cost}. Vanilla decoding and CDC use the same GPUs, numerical precision, batch size, backbone, decoding configuration, and 256-step schedule, and reported times include all surrogate and analyzer calls. Intervention frequency is the fraction of samples with at least one intervention, and successful-output throughput is the number of functional and secure outputs per unit of wall-clock time relative to vanilla decoding. MDFI reaches 86--95\% of the fs@1 of AR re-prompting (34.3 vs.\ 39.8 on CWEval and 24.7 vs.\ 26.0 on LLMSecEval+) without generating a second complete program.

\begin{table}[ht]
\centering
\caption{GradGuide inference cost on HumanEval-X C++ (164 tasks). Latency statistics are per sample; the surrogate column gives the mean (median) cost of one surrogate forward--backward pass.}
\label{tab:app_gradguide_cost}
\footnotesize
\setlength{\tabcolsep}{2.6pt}
\begin{tabular}{@{}l l r c c c c c@{}}
\toprule
Backbone & Method & Total time & Mean / p50 / p95 / max (s) & $\times$vanilla & Peak GPU & Surrogate pass & Interv.\ freq. \\
\midrule
Dream-Coder 7B & Vanilla & 693\,s & 4.23 / 4.26 / 4.62 / 5.38 & 1.00$\times$ & 15.1\,GB & -- & -- \\
 & GradGuide & 1,310\,s & 7.99 / 6.10 / 17.46 / 20.07 & 1.89$\times$ & 23.1\,GB & 49.5\,ms (47.0) & 0.66 \\
\midrule
Dream 7B & Vanilla & 685\,s & 4.18 / 4.21 / 4.57 / 5.34 & 1.00$\times$ & 14.7\,GB & -- & -- \\
 & GradGuide & 1,181\,s & 7.20 / 5.93 / 15.55 / 19.41 & 1.72$\times$ & 23.2\,GB & 52.6\,ms (48.5) & 0.84 \\
\midrule
DiffuCoder 7B & Vanilla & 682\,s & 4.16 / 4.19 / 4.54 / 5.32 & 1.00$\times$ & 14.7\,GB & -- & -- \\
 & GradGuide & 1,372\,s & 8.37 / 6.16 / 16.84 / 21.31 & 2.01$\times$ & 23.6\,GB & 48.3\,ms (45.0) & 0.66 \\
\bottomrule
\end{tabular}
\end{table}

\begin{table}[ht]
\centering
\caption{Pass@1 gain (points) of test-time-compute methods over their own vanilla baselines.}
\label{tab:app_ttc}
\small
\begin{tabular}{@{}l c c@{}}
\toprule
Method & HumanEval-X C++ $\Delta$pass@1 & MBPP-C++ $\Delta$pass@1 \\
\midrule
Best-of-$N$ ($N\!=\!2$) & +3.1 & +3.5 \\
AR+Reprompt & +4.2 & +1.0 \\
\textbf{CDC (Ours)} & \textbf{+31.1} & \textbf{+31.5} \\
\bottomrule
\end{tabular}
\end{table}

\begin{table}[ht]
\centering
\caption{MDFI inference cost with Dream-Coder 7B (vanilla$\to$MDFI).}
\label{tab:app_mdfi_cost}
\footnotesize
\begin{tabular}{@{}l r r@{}}
\toprule
Metric & CWEval & LLMSecEval+ \\
\midrule
Mean latency (s) & 12.98$\to$13.28 (+2.3\%) & 10.12$\to$10.73 (+6.1\%) \\
p50 / p95 latency (s) & 12.85$\to$12.97 / 16.73$\to$16.87 & 9.62$\to$11.00 / 11.82$\to$12.72 \\
Total benchmark time (min) & 23.4$\to$23.9 & 25.3$\to$26.8 \\
Peak GPU memory (GiB) & 64.2$\to$66.9 (+4.2\%) & 30.4$\to$31.2 (+2.6\%) \\
Analyzer calls / CPU time & 7 / 46\,ms (1 core) & 7 / 46\,ms (1 core) \\
fs@1 (\%) & 12.0$\to$\textbf{34.3} & 14.7$\to$\textbf{24.7} \\
sec@1 (\%) & 18.5$\to$\textbf{41.7} & 54.7$\to$\textbf{80.7} \\
Successful-output throughput ($\times$vanilla) & \textbf{2.79$\times$} & \textbf{1.58$\times$} \\
\bottomrule
\end{tabular}
\end{table}

\subsubsection{Ablations}
\label{sec:ablations}

\begin{figure}[t]
    \centering
    \includegraphics[width=\textwidth]{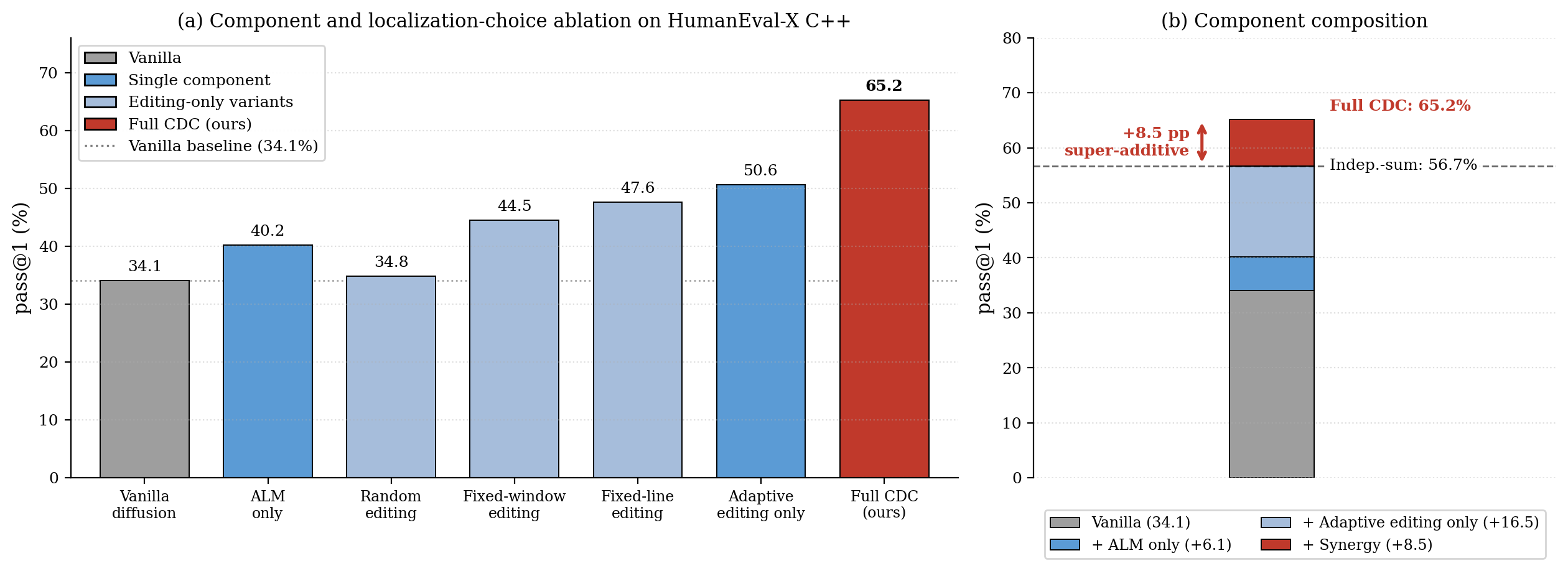}
    \caption{Component and localization-choice ablation of CDC on HumanEval-X C++ (164 tasks, Dream-Coder 7B, greedy). \textbf{(a)}~Per-configuration pass@1. \textbf{(b)}~Component composition: ALM-only +6.1\,pp and adaptive editing-only +16.5\,pp predict 56.7\% under independent action; the full operator reaches 65.2\%, a +8.5\,pp super-additive synergy.}
    \label{fig:gradguide-ablation}
\end{figure}

\paragraph{GradGuide components.}
Figure~\ref{fig:gradguide-ablation}(a) reports per-configuration pass@1 on HumanEval-X C++ (Dream-Coder 7B, greedy): vanilla 34.1, ALM-only 40.2, adaptive editing-only 50.6, full operator 65.2; random editing matches vanilla, so localization must be constraint-steered. Panel~(b) decomposes the full-operator gain: ALM-only (+6.1) and adaptive editing-only (+16.5) predict 56.7 under independent action; the full operator reaches 65.2, a \textbf{+8.5\,pp super-additive synergy} from ALM shaping the local distribution \emph{inside} the region adaptive editing reopens.

\paragraph{Correction modes.}
Table~\ref{tab:app_modes} isolates GradGuide's two correction modes on both benchmarks, keeping the backbone, surrogate, and localization fixed. Mode~A (KL-anchored proposal correction) and Mode~B (localized remasking) each improve pass@1, with Mode~B contributing more, and their combination exceeds the sum of the individual gains by 8.5 and 3.5 points on HumanEval-X C++ and MBPP-C++.

\begin{table}[ht]
\centering
\caption{Contribution of GradGuide's two correction modes (Dream-Coder 7B, greedy).}
\label{tab:app_modes}
\small
\begin{tabular}{@{}l c c c c@{}}
\toprule
Benchmark & Mode A & Mode B & pass@1 (\%) & $\Delta$ vs.\ vanilla (points) \\
\midrule
\multirow{4}{*}{HumanEval-X C++} & \ding{55} & \ding{55} & 34.1 & -- \\
 & \ding{51} & \ding{55} & 40.2 & +6.1 \\
 & \ding{55} & \ding{51} & 50.6 & +16.5 \\
 & \ding{51} & \ding{51} & \textbf{65.2} & \textbf{+31.1} \\
\midrule
\multirow{4}{*}{MBPP-C++} & \ding{55} & \ding{55} & 27.7 & -- \\
 & \ding{51} & \ding{55} & 35.5 & +7.8 \\
 & \ding{55} & \ding{51} & 47.9 & +20.2 \\
 & \ding{51} & \ding{51} & \textbf{59.2} & \textbf{+31.5} \\
\bottomrule
\end{tabular}
\end{table}

\begin{figure}[t]
    \centering
    \includegraphics[width=\textwidth]{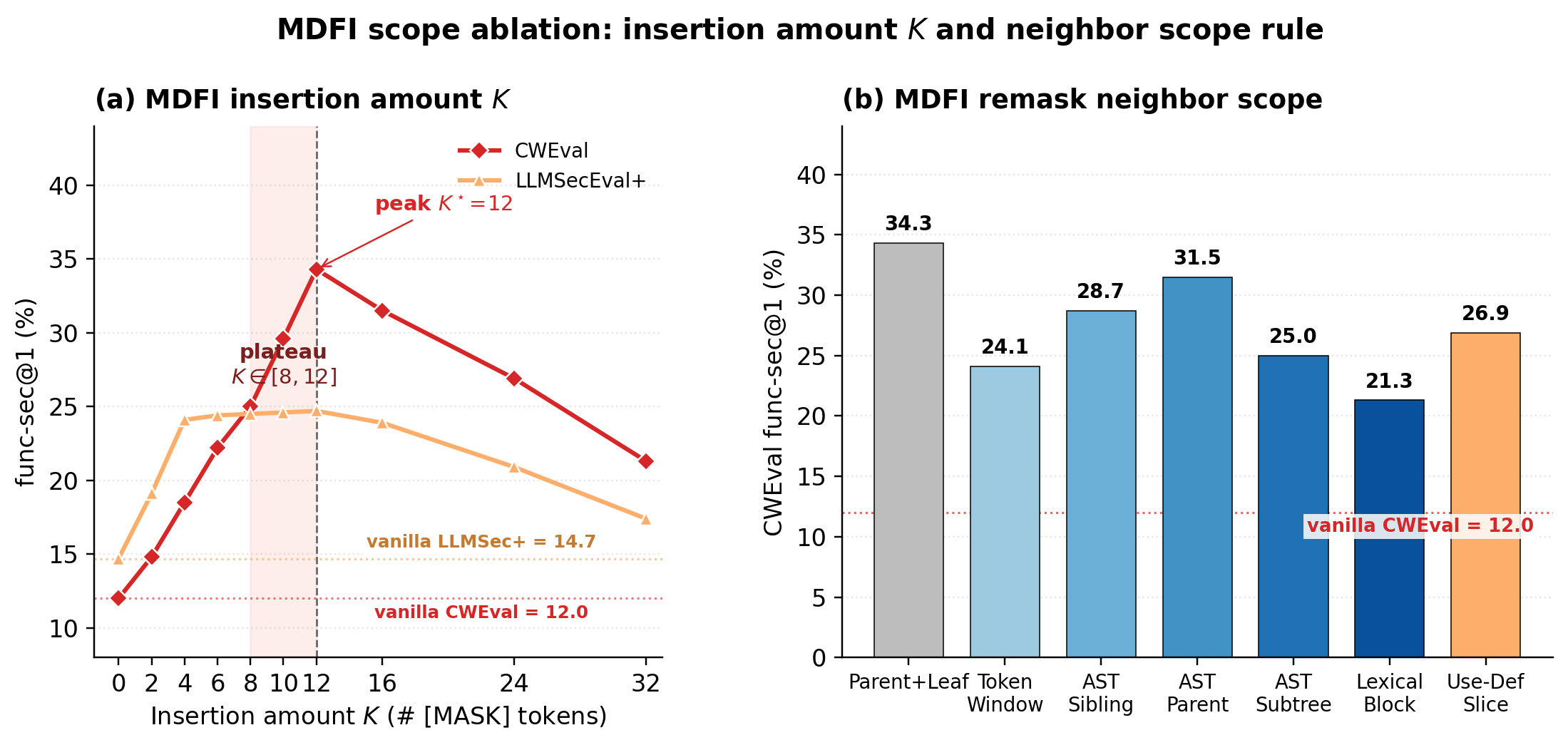}
    \caption{MDFI scope ablation. \textbf{(a)}~Insertion amount $K$ on CWEval and LLMSecEval+; func-sec@1 plateaus at $K\!\in\![8,12]$. \textbf{(b)}~Remasking neighborhood scope on CWEval; the deployed Parent$+$Leaf rule peaks at 34.3\%, beating both Token-Window and broader (Use--Def Slice) alternatives.}
    \label{fig:mdfi-ablation}
\end{figure}

\paragraph{MDFI insertion budget and neighborhood scope.}
Figure~\ref{fig:mdfi-ablation}(a): mask insertion $K$ plateaus at $K\!\in\![8,12]$ on both benchmarks (CWEval 34.3, LLMSecEval+ 24.7); $K\!>\!16$ regresses functionality. Panel~(b): the deployed Parent$+$Leaf neighborhood (34.3) beats tighter (Token-Window 24.1) and looser (Use--Def Slice 26.9) alternatives.

\subsubsection{CDC reduces correction-induced regressions}

\label{sec:functional-regression}
To complement the headline pass@$k$ rates, we characterize each method's behavior as a \emph{corrector}: how it converts vanilla outputs into corrected outputs at the per-prompt level. Following standard practice for evaluating verifier- or test-feedback-driven correctors~\cite{chen2023teachinglargelanguagemodels}, we report four complementary measurements that probe different facets of corrective behavior, summarized in Table~\ref{tab:cdc_functional_advantage}: improvement rate (fraction of base-failing prompts recovered), the raw improvement-to-regression count, net change, and constructive ratio (the share of all verdict changes that are improvements). Together they characterize whether a method recovers failures at scale, whether it does so with high precision, and whether its action is closer to a monotone corrector or to coin-flip churn.

\begin{table}[t]
\centering
\caption{Corrective effectiveness on functional benchmarks. For each method we report four complementary per-prompt measurements: \textbf{improvement rate} ($\uparrow$), the fraction of base-failing prompts the method recovers; \textbf{Imp.\,/\,Reg.} ($\uparrow$), the raw counts of improvements and regressions induced; \textbf{net} change ($\uparrow$), improvements minus regressions; and \textbf{constructive ratio} ($\uparrow$), the share of all verdict changes that are improvements (100\% corresponds to a strictly monotone corrector). For CFG-CD and CDC, the reference vanilla is Dream-Coder 7B; for AR+Reprompt, the reference is DeepSeek-Coder-Instruct-6.7B. Best per row is \textbf{bolded}. CDC is the strongest corrector on every measurement on both benchmarks: it recovers 48.0\%/43.8\% of base-failing programs (vs.\ AR+Reprompt's 28.8\%/4.4\%), delivers +46/+120 net improvements (vs.\ +2/+1), and converts 94--97\% of its verdict changes into fixes (vs.\ AR+Reprompt's near-coin-flip $\sim$53\%). CFG-CD recovers \emph{zero} failing programs on MBPP-C++.}
\label{tab:cdc_functional_advantage}
\small
\setlength{\tabcolsep}{5pt}
\renewcommand{\arraystretch}{1.05}
\begin{tabular}{l l | r r r}
\toprule
\textbf{Angle} & \textbf{Benchmark} & \textbf{CFG-CD} & \textbf{AR+Reprompt} & \textbf{CDC (Ours)} \\
\midrule
Improvement rate $\uparrow$ & HE-X C++  & 13.7\% & 28.8\% & \textbf{48.0\%} \\
 & MBPP-C++  & 0.0\% & 4.4\% & \textbf{43.8\%} \\
\midrule
Imp.\,/\,Reg.\ (counts) & HE-X C++  & 14\,/\,16 & 19\,/\,17 & \textbf{49\,/\,3} \\
 & MBPP-C++  & 0\,/\,8 & 7\,/\,6 & \textbf{124\,/\,4} \\
\midrule
Net $\uparrow$ & HE-X C++  & -2 & +2 & \textbf{+46} \\
 & MBPP-C++  & -8 & +1 & \textbf{+120} \\
\midrule
Constructive ratio $\uparrow$ & HE-X C++  & 46.7\% & 52.8\% & \textbf{94.2\%} \\
 & MBPP-C++  & 0.0\% & 53.8\% & \textbf{96.9\%} \\
\bottomrule
\end{tabular}
\end{table}

\label{sec:mdfi_regressions}
For security, Table~\ref{tab:mdfi_regressions} compares how MDFI and AR re-prompting change per-prompt verdicts on CWEval. MDFI rarely breaks programs that vanilla decoding already solves (a regression rate of at most 3.1\%, versus 21.7--29.4\% for AR re-prompting) and achieves a larger net change on every metric; all of its security and joint verdict changes are improvements, whereas AR re-prompting turns 14.7\% of its joint verdict changes into regressions.

\begin{table}[t]
\centering
\caption{Stability and corrective effectiveness on CWEval for AR re-prompting (AR) and CDC with MDFI. Regression rate is the fraction of vanilla-correct prompts that a method breaks; net change is the number of improvements minus regressions on the prompts where CDC's operator triggers; edit precision is the fraction of verdict changes that are improvements.}
\label{tab:mdfi_regressions}
\small
\setlength{\tabcolsep}{6pt}
\renewcommand{\arraystretch}{1.08}
\begin{tabular}{@{}l cc cc cc@{}}
\toprule
 & \multicolumn{2}{c}{\textbf{Regression rate (\%)} $\downarrow$} & \multicolumn{2}{c}{\textbf{Net change} $\uparrow$} & \multicolumn{2}{c}{\textbf{Edit precision (\%)} $\uparrow$} \\
\cmidrule(lr){2-3} \cmidrule(lr){4-5} \cmidrule(lr){6-7}
\textbf{Metric} & AR & CDC & AR & CDC & AR & CDC \\
\midrule
func@1 & 29.4 & \textbf{3.1} & +1 & \textbf{+19} & 51.6 & \textbf{95.2} \\
sec@1 & 23.1 & \textbf{0.0} & +29 & \textbf{+33} & 85.4 & \textbf{100.0} \\
fs@1 & 21.7 & \textbf{0.0} & +24 & \textbf{+35} & 85.3 & \textbf{100.0} \\
\bottomrule
\end{tabular}
\end{table}

\section{Details of GradGuide}
\label{app:gradguide}

This appendix provides the implementation details of the surrogate-gradient operator used in Section~\ref{subsec:operator}. GradGuide instantiates the constraint-aware operator $\mathcal{P}_{\mathcal{C}}$ for functional correctness and syntactic validity. Its goal is to provide differentiable denoising-time feedback for constraints that are ultimately evaluated by non-differentiable program oracles, such as compilation and unit-test execution.

\subsection{Surrogate Model}

Program-level correctness is non-differentiable: a candidate program either compiles and passes its tests, or it does not. GradGuide therefore trains an auxiliary surrogate $g_\phi$ ahead of time and uses it only at inference time. The diffusion model parameters $\theta$ are never updated.

At reverse step $t$, the surrogate maps the clean-state proposal $\hat{\bx}_0^{(t)} \in \Delta^{L\times |\mathcal{V}|}$ and task context $c$ to one or more continuous correctness scores,
\begin{equation}
    g_{\phi,j}(\hat{\bx}_0^{(t)},c) \in [0,1],
    \qquad j \in \{1,\dots,m\}.
\end{equation}
To make the surrogate differentiable with respect to token distributions, each soft token distribution is converted into a soft embedding,
\begin{equation}
    \mathrm{Emb}(\hat{\bx}_0^{(t)})^i
    =
    \hat{x}_0^{(t),i}\,\mathbf{E}_{\mathrm{tok}},
    \qquad i \in [L],
\end{equation}
where $\hat{x}_0^{(t),i} \in \Delta^{|\mathcal{V}|}$ is the per-position distribution (the $i$-th row of $\hat{\bx}_0^{(t)}$) and $\mathbf{E}_{\mathrm{tok}}\in\mathbb{R}^{|\mathcal{V}|\times d}$ is the surrogate input embedding matrix. The surrogate is trained using execution-derived labels, including compile success and test-pass outcomes. In our implementation, the surrogate is a LoRA-adapted Qwen2.5-Coder-1.5B backbone~\citep{hui2024qwen25coder} with a regression head on the last-token hidden state, trained on CodeContests-derived $(\text{problem}, \text{code})$ pairs labeled by $y\!=\!\mathbf{1}[\text{compiles}]\cdot(\text{test-pass ratio})\in[0,1]$ from real C++ execution; Appendix~\ref{app:surrogate_eval} gives the architecture, training data, and training protocol. The model is conditioned on a fixed judge prompt prepended to each input -- ``\emph{You are a strict unit-test judge. Given a programming problem and a solution, assign a functionality correctness score from 0 to 1: 0 = completely wrong or non-functional, 1 = fully correct and passes all intended behavior.\textbackslash n\textbackslash nProblem:\textbackslash n}'' followed by the problem text and the candidate code (token IDs at training time, soft embeddings at inference) -- so that the same prompt is used for training the regression head, for localization, and for proposal correction.

For each constraint $j$, we define the surrogate-relaxed violation
\begin{equation}
    \Delta g_j(\hat{\bx}_0^{(t)},c)
    =
    \max\!\bigl(0,\tau_j-g_{\phi,j}(\hat{\bx}_0^{(t)},c)\bigr),
\end{equation}
where $\tau_j \in (0,1]$ is the target satisfaction threshold. The aggregate surrogate violation is
\begin{equation}
    \Delta G(\hat{\bx}_0^{(t)},c)
    =
    \sum_{j=1}^m \Delta g_j(\hat{\bx}_0^{(t)},c).
\end{equation}
The surrogate functions $g_{\phi,j}$, $\mathrm{Emb}$, $\Delta g_j$, and $\Delta G$ are well-defined on any soft proposal in $\Delta^{L\times|\mathcal{V}|}$; we have written them at $\hat{\bx}_0^{(t)}$ above for clarity, and Mode~A below reuses them with its inner-loop iterate in place of $\hat{\bx}_0^{(t)}$.

\subsection{Surrogate Training and Evaluation}
\label{app:surrogate_eval}

The surrogate fine-tunes Qwen2.5-Coder-1.5B~\citep{hui2024qwen25coder} with LoRA adapters~\citep{hu2022lora} on the last 12 of its 28 decoder layers and a four-layer MLP head (dropout 0.1) on the last-token hidden state, keeping the backbone frozen. Its input consists of the judge prompt, the problem statement, and the candidate program, truncated to 2,048 tokens, and it is trained with soft-label binary cross-entropy on $y\in[0,1]$. We use a causal scorer because it reuses a pretrained code model and its last-token state attends to the entire program.

The 59K training examples are drawn exclusively from CodeContests~\citep{li2022alphacode}. Candidates are generated by Dream~\citep{ye2025dream}, Dream-Coder~\citep{xie2025dreamcoder}, and DiffuCoder~\citep{gong2025diffucoder} with $\{32,64,128,256\}$ denoising steps and temperatures $\{0,0.2,0.6,1.0\}$, supplemented by execution-verified GPT-5~\citep{openai2026gpt5} solutions for problems on which no diffusion candidate passes, and labeled by execution against the CodeContests tests. No HumanEval-X or MBPP data is used for training or model selection, and duplicate checks confirm that the training problems are disjoint from both benchmarks. We train with AdamW~\citep{loshchilov2019decoupled} (learning rate $2\times10^{-5}$, cosine schedule with 10\% warmup, weight decay 0.01, effective batch size 16, bf16) for up to five epochs and select the checkpoint with the highest validation Pearson correlation on held-out problems (epoch~3). Training takes about 1.8 GPU-hours on a single NVIDIA RTX PRO 6000 Blackwell GPU.

Table~\ref{tab:app_surrogate_eval} evaluates the surrogate before and after fine-tuning on frozen base-model candidates and on held-out CodeContests problems. MAE and MSE measure the error against $y$, Spearman $\rho$ measures ranking by test-pass ratio, and AUPRC and ECE measure discrimination and calibration for complete execution success. Guidance Utility is the fraction of same-problem candidate pairs with different outcomes that the surrogate orders correctly, and Net $\Delta$pass@1 is the change in GradGuide pass@1 (Dream-Coder 7B) when the scorer is used in the pipeline. Fine-tuning improves every metric on all three datasets and turns the pass@1 change from $-$8.5/$-$5.3 into +31.1/+31.5 points on HumanEval-X C++/MBPP-C++.

\begin{table}[ht]
\centering
\caption{Standalone evaluation of the GradGuide surrogate before fine-tuning (the Qwen2.5-Coder-1.5B backbone) and after training. Lower is better for MAE, MSE, and ECE; higher is better otherwise. N/A: not applicable.}
\label{tab:app_surrogate_eval}
\footnotesize
\setlength{\tabcolsep}{3.1pt}
\begin{tabular}{@{}l l c c c c c c c@{}}
\toprule
Scorer & Dataset & MAE & MSE & Spearman $\rho$ & AUPRC & ECE & Guidance Utility & Net $\Delta$pass@1 \\
\midrule
\multirow{3}{*}{\makecell[l]{Backbone\\(no fine-tuning)}} & HumanEval-X C++ & 0.533 & 0.342 & 0.056 & 0.277 & 0.377 & N/A & \textminus 8.5 \\
 & MBPP-C++ & 0.472 & 0.262 & 0.117 & 0.533 & 0.163 & N/A & \textminus 5.3 \\
 & Held-out CodeContests & 0.254 & 0.086 & 0.581 & 0.827 & 0.092 & 0.723 & N/A \\
\midrule
\multirow{3}{*}{\makecell[l]{Trained\\surrogate}} & HumanEval-X C++ & \textbf{0.074} & \textbf{0.022} & \textbf{0.848} & \textbf{0.946} & \textbf{0.024} & N/A & \textbf{+31.1} \\
 & MBPP-C++ & \textbf{0.111} & \textbf{0.051} & \textbf{0.837} & \textbf{0.960} & \textbf{0.030} & N/A & \textbf{+31.5} \\
 & Held-out CodeContests & \textbf{0.042} & \textbf{0.010} & \textbf{0.921} & \textbf{0.995} & \textbf{0.030} & \textbf{0.933} & N/A \\
\bottomrule
\end{tabular}
\end{table}

\subsection{Gradient-Based Localization}

GradGuide computes the gradient of the aggregate surrogate violation with respect to each soft token embedding, evaluated at the clean-state proposal $\hat{\bx}_0^{(t)}$:
\begin{equation}
    \bg_i\!\bigl(\hat{\bx}_0^{(t)},c\bigr)
    =
    \nabla_{\mathrm{Emb}(\hat{\bx}_0^{(t)})^i}
    \Delta G\!\bigl(\hat{\bx}_0^{(t)},c\bigr),
    \qquad i\in[L].
\end{equation}
The norm $\|\bg_i(\hat{\bx}_0^{(t)},c)\|_2$ measures how sensitive the predicted violation is to changes at position $i$. GradGuide combines this violation sensitivity with uncertainty signals from the base denoiser:
\begin{equation}
    a_i
    =
    \underbrace{\bigl\|\bg_i\!\bigl(\hat{\bx}_0^{(t)},c\bigr)\bigr\|_2}_{\text{violation sensitivity}}
    +
    \alpha_H H\!\bigl(\hat{x}_0^{(t),i}\bigr)
    +
    \alpha_C\!\left(1-\max_{v\in\mathcal{V}}\hat{x}_0^{(t),i}(v)\right),
\end{equation}
where $H(\hat{x}_0^{(t),i})$ is the entropy of the base proposal at position $i$, and $\alpha_H,\alpha_C\ge 0$ weight entropy and confidence terms. Importantly, the saliency is computed before any correction has been applied, so all three terms are anchored at $\hat{\bx}_0^{(t)}$. GradGuide then selects the top-$k$ saliency positions and expands them to syntactically coherent neighborhoods:
\begin{equation}
    \mathcal{S}_t
    =
    \mathrm{Expand}
    \!\left(
    \mathrm{Top}\text{-}k(a_1,\dots,a_L)
    \right).
\end{equation}
The expansion operation maps isolated token positions to coherent edit regions, such as the enclosing line, expression, block, or delimiter-balanced span. This prevents GradGuide from editing individual tokens in ways that break local syntax.

\subsection{Mode A: KL-Anchored Augmented-Lagrangian Projection}

The first correction mode shifts the clean-state proposal toward satisfying the surrogate constraints while preserving proximity to the base denoiser. Given the localized edit set $\mathcal{S}_t$, GradGuide introduces an auxiliary optimization variable $\by\in\Delta^{L\times|\mathcal{V}|}$, initialized at the clean-state proposal $\hat{\bx}_0^{(t)}$, and approximately solves
\begin{equation}
    \by_t
    \approx
    \arg\min_{\by\in\Delta^{L\times|\mathcal{V}|}}
    \mathcal{L}_{\mathrm{ALM}}
    \bigl(
    \by;\hat{\bx}_0^{(t)},\mathcal{S}_t,c
    \bigr),
\end{equation}
where
\begin{equation}
\label{eq:app_surrogate_alm}
    \mathcal{L}_{\mathrm{ALM}}
    \bigl(
    \by;\hat{\bx}_0^{(t)},\mathcal{S}_t,c
    \bigr)
    =
    D_{\mathrm{KL}}\!\left(\by\,\|\,\hat{\bx}_0^{(t)}\right)
    +
    \sum_{j=1}^{m}
    \left[
        \lambda_j\Delta g_j(\by,c)
        +
        \frac{\mu_j}{2}\Delta g_j(\by,c)^2
    \right]
    +
    \beta
    \sum_{i\notin\mathcal{S}_t}
    D_{\mathrm{KL}}\!\left(\by^i\,\|\,\hat{x}_0^{(t),i}\right).
\end{equation}
The first term is a trust region that anchors the corrected proposal to the base denoiser distribution. The second term is an augmented-Lagrangian penalty that drives surrogate violation toward zero. The final term is a locality anchor that discourages changes outside the localized edit region. In the limit of large $\beta$, the update is effectively restricted to $\mathcal{S}_t$.

In implementation, we parameterize $\by=\mathrm{softmax}(u)$ with unconstrained logits $u\in\mathbb{R}^{L\times|\mathcal{V}|}$, initialize
\begin{equation}
    u^{(0)}
    =
    \log\!\left(\hat{\bx}_0^{(t)}+\varepsilon\right),
\end{equation}
and take $K_{\mathrm{inner}}$ first-order steps on Eq.~\ref{eq:app_surrogate_alm}. The corrected proposal after these steps is denoted $\by_t$.

The multipliers and penalties are updated using an augmented-Lagrangian rule:
\begin{equation}
\label{eq:app_multipliers}
    \lambda_j
    \leftarrow
    \bigl[
    \lambda_j+\mu_j\Delta g_j(\bar{\bx}_0^{(t)},c)
    \bigr]_+,
\end{equation}
and
\begin{equation}
    \mu_j
    \leftarrow
    \begin{cases}
        \rho\mu_j,
        &
        \Delta g_j(\bar{\bx}_0^{(t)},c)
        \ge
        \vartheta \Delta g_j^{\mathrm{prev}},
        \\[2pt]
        \mu_j,
        &
        \text{otherwise},
    \end{cases}
\end{equation}
where $\rho>1$ is the penalty growth factor and $\vartheta\in(0,1)$ is a progress tolerance. The update uses the surrogate violation on the argmax-decoded intermediate program $\bar{\bx}_0^{(t)}$, rather than only the soft proposal $\by$, to reduce drift between soft surrogate satisfaction and decoded-program feasibility.

Mode A is gated by the surrogate score. If the decoded intermediate program already satisfies the surrogate threshold,
\begin{equation}
    g_\phi(\bar{\bx}_0^{(t)},c)\ge \tau_{\mathrm{alm}},
\end{equation}
then GradGuide skips the inner optimization and returns $\by_t=\hat{\bx}_0^{(t)}$. Otherwise, the KL-anchored correction is applied.

\subsection{Mode B: Constraint-Triggered Local Remasking}

The second correction mode reopens already committed tokens for re-denoising. This is necessary because, under the masked diffusion reverse kernel, once $x_t^i\neq\texttt{[MASK]}$, a proposal correction at position $i$ cannot directly rewrite the committed token. Mode B therefore remasks localized positions when the intermediate program is sufficiently decoded and still violates the target constraint.

Let
\begin{equation}
    n_{\mathrm{mask}}(\bx_t)
    =
    \bigl|
    \{i\in[L]:x_t^i=\texttt{[MASK]}\}
    \bigr|
\end{equation}
denote the number of masked positions at timestep $t$, and let $b_t$ denote the number of edits already used along the trajectory. Mode B activates when
\begin{equation}
\label{eq:app_mode_b_trigger}
    n_{\mathrm{mask}}(\bx_t)\le m_\star,
    \qquad
    b_t < B,
    \qquad
    \mathrm{Satisfies}(\bar{\bx}_0^{(t)},c)=\mathtt{false}.
\end{equation}
Here, $m_\star$ is a mask-count threshold ensuring that the partial program is sufficiently decoded for evaluation, and $B$ is the global edit budget. The predicate $\mathrm{Satisfies}$ can be an exact oracle when available, such as compilation, unit-test execution, or static analysis. When exact evaluation is not available on the partial program, GradGuide uses the surrogate condition $g_\phi(\bar{\bx}_0^{(t)},c)\ge \tau$.

When Eq.~\ref{eq:app_mode_b_trigger} is satisfied, GradGuide constructs a reopened state
\begin{equation}
\label{eq:app_remask}
    x_t^{\star,i}
    =
    \begin{cases}
        \texttt{[MASK]}, & i\in\mathcal{S}_t,\\
        x_t^i, & i\notin\mathcal{S}_t.
    \end{cases}
\end{equation}
The reverse chain then continues from $\bx_t^\star$ under the constrained reverse kernel, with Mode A active during the refill steps. The edit counter $b_t$ is incremented after the intervention.

\subsection{Composition of the Two Modes}

GradGuide uses a single surrogate signal to produce two coupled outputs at each reverse step: a corrected proposal $\by_t$ from Mode A and, when triggered, a reopened state $\bx_t^\star$ from Mode B. The constrained reverse kernel then advances the chain using the pair $(\bx_t^\star,\by_t)$. If Mode B does not activate, $\bx_t^\star=\bx_t$.

The two modes address complementary failure cases. Mode A shifts probability mass toward feasible tokens while the position remains editable through the proposal distribution. Mode B reopens tokens that have already been committed, allowing the denoiser to repair localized regions with the benefit of the surrounding program context. Both modes are scoped by the same localization set $\mathcal{S}_t$ and driven by the same surrogate gradient.

The main configurations are recovered by different choices of the projection threshold and edit budget:
\begin{align*}
    \tau_{\mathrm{alm}}=0,\; B=0
    &\quad \text{recovers the unconstrained reverse process},\\
    \tau_{\mathrm{alm}}>0,\; B=0
    &\quad \text{uses only KL-anchored proposal correction},\\
    \tau_{\mathrm{alm}}=0,\; B>0
    &\quad \text{uses only constraint-triggered remasking},\\
    \tau_{\mathrm{alm}}>0,\; B>0
    &\quad \text{uses the full GradGuide operator}.
\end{align*}
The deployed configuration uses both modes. Its per-step cost is at most one surrogate forward/backward pass, $K_{\mathrm{inner}}$ inner optimization steps when Mode A is active, and $K_{\mathrm{edit}}$ additional reverse steps when Mode B activates.

\section{Details of MDFI}
\label{app:mdfi}

This appendix provides the implementation details of the static-analysis-guided operator used in Section~\ref{subsec:operator}. MDFI instantiates the constraint-aware operator $\mathcal{P}_{\mathcal{C}}$ for security constraints. Its goal is to provide non-differentiable denoising-time feedback for properties that are characterized by discrete syntactic and dataflow patterns rather than smooth correctness scores. The base diffusion parameters $\theta$ are not modified, the analyzer carries no learnable parameters, and the same partial-program analysis pipeline is reused across all benchmarks, languages, and decoding configurations: MDFI is therefore training-free.

\subsection{Partial Program Representation}
\label{app:mdfi-pcpg}

At each fired checkpoint, the analyzer builds a structural representation $\mathcal{G}_t$ from the decoded clean-state proposal $\bar{\bx}_0^{(t)}$ defined in Eq.~\ref{eq:decode_intermediate}. The representation is a partial program graph that combines an abstract syntax tree with a dataflow approximation: $\mathcal{G}_t = (V,\,E^{\mathrm{AST}},\,E^{\mathrm{CFG}},\,E^{\mathrm{DFG}},\,\ell)$, where $V$ are program nodes (statements, expressions, identifiers), $E^{\mathrm{AST}}$ encodes parent--child structure, $E^{\mathrm{CFG}}$ encodes control flow between adjacent basic blocks within each function, $E^{\mathrm{DFG}}$ encodes intra-procedural dataflow between defining and using occurrences of each program identifier, and $\ell$ labels each node with its lexical class.

Tokens in $\bx_t$ that remain masked at checkpoint time are not carried into the parser literally; instead, each mask is rewritten into a fresh placeholder identifier of the form \texttt{\_\_hole\_<i>\_\_} before parsing. This rewrite preserves the lexical class of the position---an identifier slot remains an identifier, a literal slot remains a literal---and lets the partial parser produce a well-formed graph even when a non-trivial fraction of the program is still masked. The analyzer is invoked at a sparse schedule of checkpoints (Section~\ref{app:mdfi-schedule}), gated by a minimum committed-fraction so that the rewrite is only triggered when enough of $\bar{\bx}_0^{(t)}$ has been decoded for the resulting graph to be informative.

\subsection{Vulnerability Detection on Partial Programs}
\label{app:mdfi-detection}

The detector inspects $\mathcal{G}_t$ for two complementary symptoms of insecurity that together cover both fully decoded and partially masked vulnerable shapes.

\textbf{Dataflow witnesses.} For each known unsafe sink class, the detector performs a bounded breadth-first search on $E^{\mathrm{DFG}}$ from candidate \emph{source} nodes (locations in $\mathcal{G}_t$ that introduce data outside the program's trust boundary, such as user input or environment data) toward candidate \emph{sink} nodes (locations that consume data in a security-sensitive way, such as command execution or unparameterized query construction). The search halts whenever it crosses a sanitizer node and yields a witness whenever it terminates at a sink without ever crossing one. The result is a path of nodes in $\mathcal{G}_t$ that demonstrates the vulnerable flow.

\textbf{Structural witnesses.} For partial programs in which the dangerous identifier is still masked or the dataflow chain has not yet crystallized, the detector also matches local AST shapes that are unsafe regardless of dataflow: a sink callsite whose argument shape bypasses a required guard, an insecure construction template, or a missing structural neighbor that the safe usage requires. These shape matches fire on subgraphs of $\mathcal{G}_t$ rather than on dataflow paths and are essential for catching vulnerabilities at early checkpoints when only the coarse program skeleton is committed.

In both cases, each detection produces a witness $w_k=(n_k,\tau_k,h_k)$ with the offending node $n_k\in V$, the correction type $\tau_k\in\{\mathrm{sub},\mathrm{ins}\}$, and a structured remediation hint $h_k$ identifying the violated property and the recommended safe pattern. The witnesses across the two detection paths are combined into a single set $\mathcal{W}_t = \{w_k\}_{k=1}^{K_t}$ that populates the violation vector $\boldsymbol{\nu}_t$ and structured feedback $r_t$ in Eq.~\ref{eq:constraint_localization}.

\subsection{Analyzer Rules and Choice of Static Analysis}
\label{app:mdfi-analyzer}

The analyzer reuses existing declarative detection rules and intermediate representations from three general-purpose static analyzers, CodeQL\footnote{\url{https://codeql.github.com}}, Semgrep\footnote{\url{https://semgrep.dev}}, and Bandit\footnote{\url{https://github.com/PyCQA/bandit}}, without additional hand-crafted rules; the partial-program layer described above adapts them to masked, incomplete programs and maps each finding to an editable token region. We use static analysis rather than a learned detector because the available labeled data, 258 prompts across CWEval (108) and LLMSecEval+ (150), are too few to train a reliable multilingual detector and would leak evaluation data. A false positive may trigger an unnecessary correction, which mainly adds overhead. As Table~\ref{tab:app_analyzers} shows, substituting the off-the-shelf tools for our analyzer on LLMSecEval+ improves fs@1 by at most 2.6 points at 36--288\% additional latency (CodeQL times out), whereas our analyzer improves fs@1 by 10.0 points at 6.1\% additional latency.

\begin{table}[ht]
\centering
\caption{MDFI with different static analyzers on LLMSecEval+ (Dream-Coder 7B).}
\label{tab:app_analyzers}
\small
\begin{tabular}{@{}l c c c c@{}}
\toprule
Analyzer & sec@1 (\%) & fs@1 (\%) & $\Delta$fs@1 vs.\ vanilla & Latency vs.\ vanilla \\
\midrule
None (vanilla, no MDFI) & 54.7 & 14.7 & -- & -- \\
Bandit & 62.7 & 17.3 & +2.6 & +36\% \\
Semgrep & 62.7 & 17.3 & +2.6 & +288\% \\
CodeQL & 54.7 & 14.7 & 0.0 & timed out \\
\textbf{Our partial-program analyzer} & \textbf{80.7} & \textbf{24.7} & \textbf{+10.0} & \textbf{+6.1\%} \\
\bottomrule
\end{tabular}
\end{table}

\subsection{Localization on the Program Graph}
\label{app:mdfi-localization}

For each witness $w_k$, the localization map $\mathcal{M}^{\mathrm{MDFI}}$ lifts the offending node $n_k$ to a coherent token region by walking $\mathcal{G}_t$ along its surrounding AST and dataflow neighborhood. The neighborhood includes the smallest enclosing statement node, the immediately neighboring AST nodes around $n_k$, and the dataflow-adjacent nodes that define identifiers used by $n_k$ within a small dataflow radius. The set of structurally adjacent nodes $\mathcal{N}(n_k;\mathcal{G}_t)$ is then projected back to token positions:
\begin{equation}
\label{eq:app-mdfi-loc-perwit}
    N(w_k)
    \;=\;
    \mathrm{Tok}\!\bigl(\mathcal{N}(n_k;\mathcal{G}_t)\bigr)
    \;\subseteq\;
    [L].
\end{equation}
The full editable region is the union over witnesses, capped by a token budget $B$ that bounds the total fraction of the program that may be revised at any single checkpoint:
\begin{equation}
\label{eq:app-mdfi-loc}
    \mathcal{S}_t
    \;=\;
    \mathrm{TopBudget}_{B}\!\Bigl(\bigcup_{k=1}^{K_t} N(w_k)\Bigr)
    \;\subseteq\;
    [L].
\end{equation}
When $\bigl|\bigcup_{k} N(w_k)\bigr|>B$, $\mathrm{TopBudget}_B$ retains the witnesses with the highest analyzer confidence first; ties are broken at the granularity of AST statements so that the resulting localization preserves syntactic coherence.

\subsection{Substitute Remasking}
\label{app:mdfi-sub}

For witnesses with $\tau_k=\mathrm{sub}$, the offending construct is already present in the partial program in the wrong form---for instance, an unsafe API call, an unsafe constructor, or a hard-coded credential. The operator opens these positions for re-denoising by setting them back to the mask token. Let
\begin{equation}
    \mathcal{S}_t^{\mathrm{sub}}
    \;=\;
    \mathcal{S}_t \;\cap\; \!\!\bigcup_{k:\tau_k=\mathrm{sub}}\!\! N(w_k).
\end{equation}
The substitute operation produces
\begin{equation}
\label{eq:app-mdfi-remask}
    x_t^{\star,i}
    \;=\;
    \begin{cases}
        \texttt{[MASK]}, & i\in\mathcal{S}_t^{\mathrm{sub}},\\
        x_t^i, & i\notin\mathcal{S}_t^{\mathrm{sub}}.
    \end{cases}
\end{equation}
This is the same form as the standard partial-mask state defined in Section~\ref{subsec:constrained_denoising}, with the localization set anchored on a witness node rather than on a heuristic score.

\subsection{Mask Insertion}
\label{app:mdfi-ins}

For witnesses with $\tau_k=\mathrm{ins}$, the bug is the \emph{absence} of a needed construct---for example, a missing input-validation guard, a missing length check, or a missing exception handler. Re-denoising existing tokens cannot resolve these failures because the tokens that should be there do not yet exist in $\bx_t$. The operator instead allocates $K$ fresh mask positions adjacent to a structurally chosen anchor in $N(w_k)$ and splices them into the partial-mask state:
\begin{equation}
\label{eq:app-mdfi-insert}
    \bx_t^{\star}
    \;\leftarrow\;
    \mathrm{Insert}_{K}\!\bigl(\bx_t^{\star},\;\mathrm{anchor}(w_k)\bigr),
    \qquad k:\tau_k=\mathrm{ins}.
\end{equation}
The anchor is chosen so that the inserted region admits a syntactically coherent infill---typically immediately before the offending sink node, immediately after a relevant assignment, or at the start of the offending block. The insertion extends the sequence length by $K$ and right-shifts the downstream committed tokens to preserve syntactic context. The localization set $\mathcal{S}_t^{\mathrm{sub}}$ used by Eq.~\ref{eq:app-mdfi-remask} is updated to reference the post-insertion sequence whenever both substitute and insert witnesses are present in $\mathcal{W}_t$.

\subsection{Pre-Allocated Prompt Buffer}
\label{app:mdfi-buffer}

To prevent the next denoiser pass from regenerating the same vulnerable construct under unchanged conditioning, MDFI augments the conditioning context $c$ with the analyzer's remediation hints. At the start of every trajectory we pre-allocate a contiguous mask buffer of fixed length $B_p$ inside the prompt context, immediately following the task description. This buffer is part of the model's input but is flagged as conditioning rather than as part of the output sequence; the constrained reverse kernel never samples into it, and base-model attention over fixed prompt positions is unaffected by it.

At each fired checkpoint, the structured hints $\{h_k\}$ in $r_t$ are tokenized into a natural-language remediation message $\rho_t(r_t)$ that names the violated property and the recommended safe pattern, and $\rho_t(r_t)$ is written into the buffer slots, displacing leading mask tokens. When multiple checkpoints fire along a trajectory, the most recent message overwrites any previously injected message in the same buffer:
\begin{equation}
\label{eq:app-mdfi-buffer}
    c^{\star}
    \;=\;
    \mathrm{Write}\!\bigl(c,\;\mathrm{Buffer}_{B_p},\;\rho_t(r_t)\bigr).
\end{equation}
Because $|c^{\star}|=|c|$ by construction, no token positions inside the program $\bx_t^{\star}$ are displaced by feedback injection. When the message is shorter than $B_p$, trailing buffer slots remain mask tokens and the model can still infer ``no further hint'' from that region.

\subsection{Checkpoint Schedule and Composition}
\label{app:mdfi-schedule}

The MDFI operator is invoked at a predetermined schedule of denoising checkpoints
\begin{equation}
    \mathcal{T}^{\mathrm{ck}}\;\subseteq\;\{1,\dots,T\},
\end{equation}
gated by a minimum committed-fraction $\rho_{\min}\!\in\![0,1)$:
\begin{equation}
\label{eq:app-mdfi-trigger}
    \frac{\bigl|\{i\in[L]:x_t^i\neq\texttt{[MASK]}\}\bigr|}{L}
    \;\ge\;
    \rho_{\min}.
\end{equation}
This guarantees that the analyzer is consulted only once enough of $\bar{\bx}_0^{(t)}$ has been decoded for the partial program graph $\mathcal{G}_t$ to be informative. A global intervention budget $B_{\mathrm{int}}$ caps the total number of fired checkpoints along any single trajectory; once exhausted, the operator collapses to the identity for the rest of the chain.

When a checkpoint at step $t$ is active, the operator combines the substitute remask of Eq.~\ref{eq:app-mdfi-remask}, the mask insertion of Eq.~\ref{eq:app-mdfi-insert}, and the buffer write of Eq.~\ref{eq:app-mdfi-buffer} into a single output:
\begin{equation}
\label{eq:app-mdfi-step}
    \bigl(\bx_t^{\star},\,c^{\star},\,\mathcal{S}_t\bigr)
    \;=\;
    \mathcal{P}_{\mathcal{C}}^{\mathrm{MDFI}}\!\bigl(\hat{\bx}_0^{(t)},\,\mathcal{S}_t,\,r_t,\,c\bigr).
\end{equation}
The chain then advances from the modified state and conditioning under the constrained reverse kernel of Eq.~\ref{eq:constrained_reverse_update}: $\bx_{t-1}\sim p_\theta^{\mathcal{C}}(\,\cdot\mid\bx_t^{\star},\hat{\bx}_0^{(t)},\mathcal{S}_t;\,c^{\star})$. Crucially, MDFI does not run an inner re-denoising loop on the reopened region; it modifies the state and conditioning in place and proceeds directly to the next reverse step. As a result, the freshly masked positions are refilled by the remaining $t$ standard transitions under feedback-aware conditioning, and the total number of reverse transitions along the trajectory is identical to vanilla diffusion.

\end{document}